\documentclass[nohyperref]{article}

\usepackage{microtype}
\usepackage{graphicx}
\usepackage{subfigure}
\usepackage{booktabs} 
\usepackage{adjustbox}

\usepackage{algorithm, algorithmic}

\usepackage[nonatbib,preprint]{neurips_2026}

\usepackage[utf8]{inputenc} 
\usepackage[T1]{fontenc}    
\usepackage[pagebackref,breaklinks,colorlinks]{hyperref}

\usepackage{graphicx}
\usepackage{amsmath}
\usepackage{amssymb}
\usepackage{booktabs}
\usepackage[utf8]{inputenc} 

\usepackage{adjustbox}

\usepackage{natbib}

\usepackage[nameinlink]{cleveref}
\crefname{section}{Sec.}{Secs.}
\Crefname{section}{Section}{Sections}
\Crefname{table}{Table}{Tables}
\crefname{table}{Tab.}{Tabs.}

\usepackage{pifont}

\usepackage{url}

\usepackage{tikz}
\usepackage{comment}
\usepackage{amsfonts}       
\usepackage{color}

\usepackage{amsmath}
\usepackage{amssymb}
\usepackage{bbm, dsfont}
\usepackage{mathtools}
\usepackage{amsthm}
\usepackage[normalem]{ulem}
\usepackage{xcolor,colortbl}
\usepackage{multirow}
\usepackage{cutwin}
\usepackage{arydshln}
\usepackage{enumitem}
\usepackage{wrapfig}
\usepackage{microtype}
\usepackage{graphicx}
\usepackage{bm}
\usepackage{eqparbox}
\usepackage{makecell}
\usepackage{threeparttable}
\usepackage{wrapfig}

\newcommand{\eat}[1]{}

\definecolor{Red}{rgb}{0.6,0,0}
\definecolor{Blue}{rgb}{0,0,0.8}
\definecolor{Green}{rgb}{0,0.7,0.3}
\definecolor{airforceblue}{rgb}{0.36, 0.54, 0.66}
\definecolor{ao(english)}{rgb}{0.0, 0.5, 0.0}
\definecolor{azure(colorwheel)}{rgb}{0.0, 0.5, 1.0}
\definecolor{crimson}{rgb}{0.86, 0.08, 0.24}
\definecolor{darkcerulean}{rgb}{0.03, 0.27, 0.49}
\definecolor{cobalt}{rgb}{0.0, 0.28, 0.67}
\definecolor{rosegold}{rgb}{0.72, 0.43, 0.47}
\definecolor{orange-red}{rgb}{1.0, 0.27, 0.0}
\definecolor{mountainmeadow}{rgb}{0.19, 0.73, 0.56}
\definecolor{malachite}{rgb}{0.04, 0.85, 0.32}
\definecolor{darkblue}{rgb}{0.0, 0.0, 0.55}
\definecolor{customblue}{rgb}{0.2, 0.35, 0.8}
\definecolor{gg}{gray}{0.92}
\newcolumntype{a}{>{\columncolor{gg}}c}

\definecolor{gg}{gray}{0.9}

\hypersetup{colorlinks=true}
\hypersetup{linktoc=all}
\hypersetup{citecolor=customblue}
\hypersetup{linkcolor=crimson}
\hypersetup{urlcolor=darkblue}
\usepackage[all]{hypcap}

\creflabelformat{equation}{#2\textup{#1}#3}

\usepackage{svg}

\title{Mixtures of SubExperts \\ for Large Language Continual Learning}

\author{
    Haeyong Kang$^{*}$, Hee Suk Yoon, Dahua Feng, and Chang D. Yoo \\
    Korea Advanced Institute of Science and Technology (KAIST) \\
    {\small \texttt{haeyong.kang@kaist.ac.kr}}
}

\begin{document}

\maketitle

\begin{abstract}
Enabling lifelong learning in LLMs demands resolving the stability-plasticity dilemma (i.e., models must incorporate new knowledge without overwriting prior representations) while maintaining scalability under bounded parameter growth. Existing PEFT methods fail to satisfy this triad; shared-parameter approaches suffer from catastrophic interference, while task-isolated expansions preclude knowledge transfer and scale linearly. We propose \textit{Mixtures of SubExperts (MoSEs)}, a modular and sparse framework that factorizes model capacity into reusable, compositional primitives. MoSEs augment transformer layers with lightweight SubExperts and a learned sub-routing function that dynamically selects and composes a sparse subset of modules conditioned on task inputs. This induces a structured decomposition of the parameter space where knowledge is localized yet accessible, mitigating interference while preserving reuse. Specifically, MoSEs balance the dilemma via three pillars: (i) \textbf{\textit{stability}} by isolating knowledge within sparsely activated modules, (ii) \textbf{\textit{plasticity}} through routing-driven recombination and selective expansion, and (iii) \textbf{\textit{scalability}} via sublinear growth in effective capacity. Notably, the routing mechanism enables compositional generalization, allowing new tasks to be represented as combinations of previously acquired sub-functions. We empirically validate MoSEs on TRACE and SuperNI, showing reduced forgetting, improved forward transfer, and better parameter efficiency over strong PEFT baselines. MoSEs establish a new Pareto frontier, achieving state-of-the-art performance while maintaining strict parameter budgets. Our results suggest that modular sparsity and compositional routing are key inductive biases for building foundation models that continually learn without saturation.
\end{abstract}

\section{Introduction}

Large Language Models (LLMs) have significantly advanced the state of natural language processing (NLP), powering systems that perform tasks such as summarization, question answering, dialogue, translation, and reasoning \cite{brown2020language, openai2023gpt4, touvron2023llama}. These models are typically pretrained on massive corpora using self-supervised learning and subsequently fine-tuned for specific downstream applications. However, their training paradigm is inherently static: once deployed, LLMs cannot easily incorporate new knowledge or adapt to evolving domains without costly retraining or fine-tuning on large datasets.

In real-world scenarios, where data continuously arrives, and user demands shift over time, models must be updated efficiently without degrading past performance. \emph{Continual Learning} (CL) \cite{ThrunS1995, rusu2016progressive, zenke2017} addresses this challenge by enabling models to acquire new tasks sequentially while retaining previously learned knowledge \cite{parisi2019continual}. Unfortunately, deep neural networks - especially those with shared parameters - suffer from \emph{catastrophic forgetting} when fine-tuned sequentially \cite{mccloskey1989catastrophic, kirkpatrick2017overcoming}, as gradients for new tasks often override information critical for earlier ones. Although CL has been extensively studied in smaller-scale settings such as image classification, adapting it to large-scale language models remains an open problem. LLMs introduce unique challenges for CL, including high memory requirements, difficulty in task boundary detection, and interference due to overparameterization.

Several conventional strategies have been proposed to mitigate forgetting, but most struggle with scalability and efficiency. Replay-based methods \cite{lopez2017gradient, chaudhry2019tiny} maintain exemplar buffers from previous tasks and interleave them during training. However, storing real data raises privacy and compliance concerns, particularly in sensitive domains such as healthcare or finance. Moreover, replay buffers grow linearly with the number of tasks, limiting their practicality. Regularization-based methods, such as Elastic Weight Consolidation (EWC) \cite{kirkpatrick2017overcoming} and Synaptic Intelligence (SI) \cite{zenke2017}, and Architecture-based methods, such as Supermasks (SupSup) \cite{wortsman2020supermasks} and Winning SubNetworks (WSN) \cite{kang2022forget}, restrict updates to parameters deemed important for past tasks. These methods, while effective in small models - convolutional neural networks (CNNs), often underperform on high-dimensional parameter spaces like transformers due to limited plasticity and reliance on heuristic importance metrics.

\begin{figure*}[t!]
    \vspace{-10pt}
    \centering
    \setlength{\tabcolsep}{-3.pt}{%
    \includegraphics[height=4.6cm, trim={1.cm 1.cm 1.0cm 1.0cm},clip]{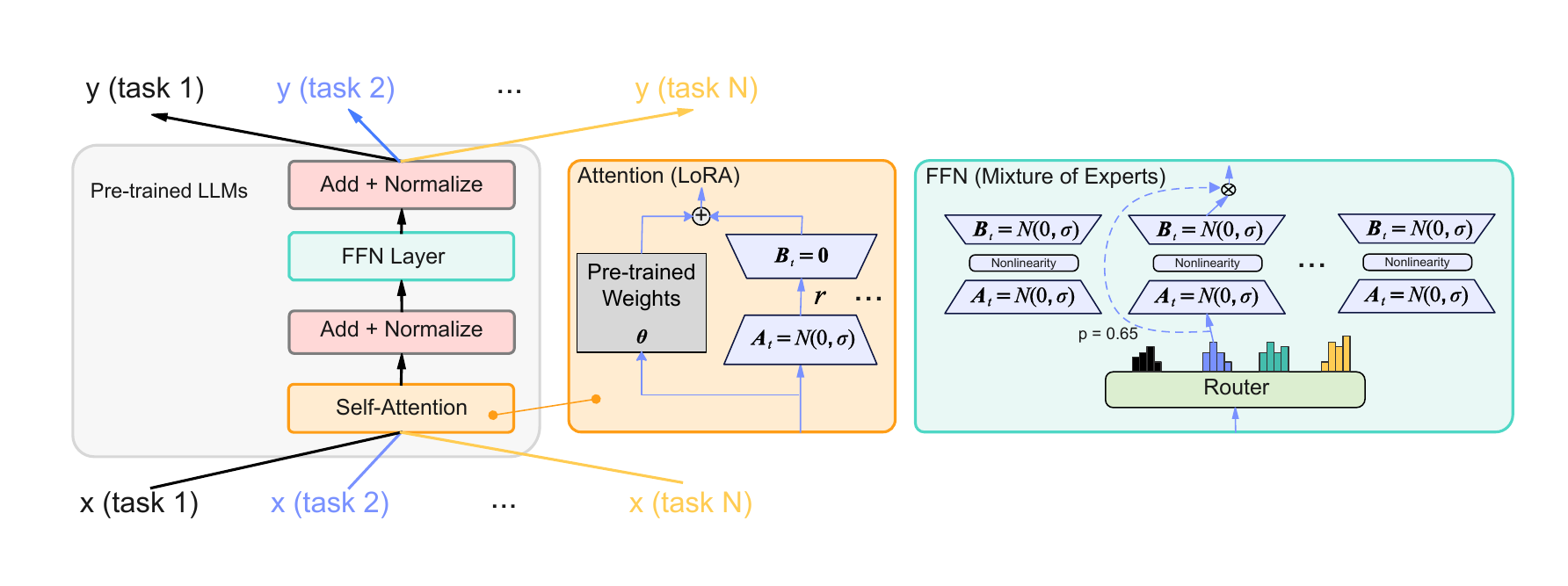}
    }
    \vspace{-15pt}
    \caption{\small \textbf{Prior PEFT frameworks for continual language modeling.} A pre-trained transformer (left) is fine-tuned to handle a stream of sequential tasks without retraining the frozen backbone. Two parameter-efficient fine-tuning (PEFT) approaches are widely used: 1) \textbf{LoRA} (middle) augments the frozen attention weights $\theta$ with a low-rank update, 2) \textbf{Mixture of Experts (MoE)} (right) replaces the feed-forward network (FFN) layer with a router-controlled bank of expert sub-networks, activating a sparse subset per input token. $\bm{A}, \bm{B}$ are the learnable parameters, initialized respectively.}
    \label{fig:peft_outline}
    \vspace{-10pt}
\end{figure*}

In transformer-based LLM fields, Parameter-efficient fine-tuning (PEFT, \Cref{fig:peft_outline}) offers an attractive alternative by modifying only a small subset of model parameters for each task, reducing computational overhead and allowing quick deployment \cite{houlsby2019parameter, hu2021lora, li2021prefix}. Adapter modules, LoRA (Low-Rank Adaptation), and prefix tuning have shown strong performance in multitask and transfer learning scenarios. However, naive application of PEFT to CL still leads to severe forgetting, as task-specific parameters can interfere without proper isolation or coordination. Another promising direction is the use of \emph{Mixture of Experts} (MoE) architectures \citep{shazeer2017outrageously, lepikhin2020gshard, fedus2022switch}, which introduce sparse routing to different expert subnetworks, increasing model capacity without proportionally increasing inference cost. While MoEs allow for modular computation and dynamic routing, they have not yet been successfully adapted for large language continual learning. Without a mechanism to prevent expert overlap or drift, MoE-based models still experience performance degradation in CL settings \citep{li2025theory}.

To overcome these limitations, we propose a novel method for continual learning in LLMs, referred to as \textbf{\emph{M}}ixture of \textbf{\emph{S}}ub\textbf{\emph{E}}xperts (\textbf{MoSEs}) for parameter-efficient fine-tuning. Our method is designed to integrate the strengths of expert-based sparse computation with the efficiency of PEFT methods, enabling scalable, robust, and interference-rare LLM continual learning across tasks.

Our contributions are summarized as follows:
\begin{itemize}[leftmargin=3.5mm]
    \item We introduce a sparsely-routed Mixture of SubExperts (MoSEs) framework in which sparse experts are selected and their sparse parameters are adaptively overlapped across tasks, thus maximizing the model plasticity and scalability.
    \item Adaptively by selecting task-specific sub-experts with prompt embedding in sparse attention layers, our MoSEs achieve minimal catastrophic forgetting while reducing computational overhead.
    \item We evaluate MoSEs on the TRACE and SuperNI benchmark datasets, which include a diverse set of language tasks, demonstrating that MoSEs are the SOTA method in terms of performance, forgetting, and parameter efficiency.
\end{itemize}

\section{Related Work}\label{sec:related}
To adapt pretrained language models efficiently, parameter-efficient fine-tuning (PEFT) methods have emerged as practical alternatives to full model updates. These include Adapters~\cite{houlsby2019parameter}, Prefix-Tuning~\cite{li2021prefix}, and Low-Rank Adaptation (LoRA)~\cite{hu2021lora}. These methods introduce small, trainable modules while keeping most of the backbone frozen. LoRA, for example, inserts low-rank matrices into attention layers, achieving performance comparable to full fine-tuning with significantly fewer trainable parameters. While PEFT methods excel at reducing resource demands and enabling multi-task deployment, they are not a suitable solution for sequential learning, as all parameters are reused.  In Continual Learning (CL) settings, updating adapters or LoRA weights across tasks can still result in parameter interference unless task-specific modules are carefully isolated. Methods such as AdapterFusion and Prompt Tuning, i.e., L2P~\cite{wang2022learning}, DualPrompt~\cite{wang2022dualprompt}, HidePrompt~\cite{wang2023HiDe}, and NoRGa~\cite{le2025NoRGa}, offer some modularity, but fail to scale to long task sequences or handle forgetting explicitly. Recent works have combined PEFT with continual learning~\citep{zhang2023adaptive}, but often lack routing mechanisms that control the sharing of experts or protection between tasks to maximize the model plasticity and minimize catastrophic interference.

To address the limitations of routing mechanisms, Mixture-of-Experts (MoE) models were introduced to decouple model capacity from computational cost by activating only a subset of experts per input~\cite{shazeer2017outrageously, lepikhin2020gshard, fedus2022switch}, demonstrating remarkable scalability in models such as GLaM~\cite{du2022glam} and Switch Transformers~\cite{fedus2022switch}. While the sparse routing, conditional computation of LoRAMoE~\cite{dou2024loramoe} offers natural potential for CL, MoEs trained in multi-task settings suffer from entangled expert specialization, making them susceptible to catastrophic forgetting and lacking post-training assignment flexibility. Similarly, recent advances in Parameter-Efficient Fine-Tuning (PEFT) have attempted to mitigate forgetting, yet struggle with fundamental structural hurdles; SeqLoRA and LoRA~\cite{hu2021lora} lack explicit isolation, while more advanced methods like O-LoRA~\cite{wang2023olora} and InfLoRA~\cite{liang2024inflora} utilize orthogonal constraints and informative parameter identification to protect prior knowledge. Furthermore, SAPT~\cite{zhao2024sapt} aligns parameter-efficient tuning with task selection through a shared attention mechanism, and TASL ~\cite{feng2024tasl} utilizes task skill localization to identify a small subset of parameters responsible for task-specific performance. Despite the introduction of gated modulation in GainLoRA~\cite{liang2025gated}, hierarchical decomposition in HiDeLoRA~\cite{wang2025hide}, and the tree-structured parameter branching of TreeLoRA~\cite{qian2025treelora} to manage task dependencies, these approaches still operate within shared or incrementally constrained parameter spaces, failing to provide the robust structural isolation and scalable capacity needed for task sequences. Addressing these limitations, our work integrates PEFT with interference-minimized sparse routing, enabling adaptive parameter sharing across SubExperts specialized for each task to minimize forgetting while maximizing parameter efficiency.

\section{Prerequisites}
We begin by reviewing conventional fine-tuning approaches, including Low-Rank Adaptation (LoRA)~\citep{hu2021lora} and Mixture-of-Experts (MoE)~\citep{shazeer2017outrageously}, and discuss the key challenges these methods present for continual learning (CL) in LLMs.

\subsection{Preliminaries}
\paragraph{Problem Statement.}
Continual Learning (CL) involves training deep neural networks on a time-varying data stream represented as a sequence of tasks, $\mathcal{D} = \{\mathcal{D}_1, \cdots, \mathcal{D}_\mathcal{T}\}$. Each task $\mathcal{D}_t = \{(\bm{x}_i^t, \bm{y}_i^t)\}_{i=1}^{n_t}$ consists of $n_t$ data points, where $\bm{x}_i^t \in \mathcal{X}_t$ denotes an input sample, and $\bm{y}_i^t \in \mathcal{Y}_t$ is the corresponding label. Upon arrival of the $t$-th task $\mathcal{D}_t$, the model $f_{\bm{\theta}}$ is updated using only the current task data, as previous data $\mathcal{D}_{<t}$ from earlier tasks are no longer accessible. This work focuses primarily on task-agnostic continual language learning (TAG-CLL), a challenging CL setting, where the task identity is available during training as the sequence index but is not provided at inference time. The learning objective in this setting is formally following $\max_{\bm{\theta}} \sum_{t=1}^{T}\sum_{(\bm{x}_i^t,\bm{y}_i^t) \in \mathcal{D}_t} \log p_{\bm{\theta}} (\bm{y}^t_i | \bm{x}^t_i)$.

\subsection{Low-Rank Adaptation (LoRA)}
Low-Rank Adaptation (LoRA)~\citep{hu2021lora}, stated in \Cref{fig:peft_outline}, is a PEFT method tailored to adapt large pre-trained models, such as Transformers, by introducing trainable low-rank matrices into selected layers while keeping the original pretrained weights $\bm{\theta}$ frozen at all attention layers. This strategy has proven effective in continual learning scenarios involving large language models (LLMs), where both computational efficiency and mitigation of catastrophic forgetting are essential. Rather than updating the full pre-trained weight matrix $\bm{\theta} \in \mathbb{R}^{d \times d}$ during training, LoRA re-parameterizes it as:
\begin{equation}
\bm{\theta}' = \bm{\theta} + \Delta \bm{\theta}, \quad \text{where} \quad \Delta \bm{\theta} = \bm{B}\bm{A},
\end{equation}
where $\bm{A} \in \mathbb{R}^{r \times d}$ and $\bm{B} \in \mathbb{R}^{d \times r}$ are low-rank matrices with rank $r \ll d$. The base matrix $\bm{\theta}$ is frozen and only low-rank matrices $\bm{A}$ and $\bm{B}$ are trained. The update $\Delta \bm{\theta}$ is scaled by a factor $\alpha / r$ to regulate its magnitude. During inference, the low-rank update can either be merged into $\bm{\theta}$ for efficient execution or retained as a modular adapter to allow flexible task switching.

LoRA offers several advantages for continual learning in LLMs by drastically reducing the number of trainable parameters, enabling scalable multi-task training, and preventing catastrophic forgetting through a frozen base model that preserves prior knowledge. Its modular design further supports task-specific adaptation, transfer learning, and efficient storage or communication since only lightweight adapter weights need to be saved or transmitted. However, LoRA also has practical limitations: as tasks accumulate, the growing collection of adapters can become a memory and storage burden; unmerged adapters require task-specific weight loading at inference time, complicating deployment; and reusing adapters across tasks may introduce catastrphic interference unless carefully managed.

\subsection{Mixture-of-Experts (MoE)}
Mixture-of-Experts (MoE), shown in \Cref{fig:peft_outline}, is a modular architecture that enhances neural networks by introducing a set of parallel expert subnetworks, combined with a routing mechanism that selects a subset of these experts for each input~\cite{shazeer2017outrageously} at all feed-forward network (FFN) layers. MoE introduces a set of $N$ expert networks and a routing function that selects a sparse subset of $k \ll N$ experts for each input, allowing conditional computation. For input $\bm{x}$, the output is:
\begin{equation}
\text{MoE}(\bm{x}) = \sum_{j \in \text{TopK}(R(\bm{x}))}^k R_j(\bm{x}) \cdot E_j(\bm{x}),
\label{eq:moe}
\end{equation}
where $R_j(\bm{x})$ is the dense routing weight and $E_j$ is the $j$-th dense expert.
This approach has gained significant traction in scaling LLMs efficiently and is increasingly adopted in continual learning scenarios. 

In the MoE, each expert is a feedforward network, and a trainable gating function determines which $k$ out of $N$ experts should process a given input. The use of sparse gating - activating only a small subset of experts - ensures that the computational overhead remains low, even when the number of experts is large. However, MoE methods also face practical challenges: some experts may be overused while others remain idle, reducing learning efficiency and generalization; the gating mechanism can become unstable or collapse without careful regularization, which is especially problematic in continual learning; deployment becomes more complex due to task-specific routing and expert configurations; and as tasks accumulate, managing and storing these configurations also becomes a significant scalability bottleneck.

\section{Our Mixtures of SubExperts (MoSEs)}
\begin{figure*}[t]
    \vspace{-10pt}
    \centering
    \includegraphics[height=5.6cm, trim={0.8cm 0.8cm 0.8cm 0.8cm},clip]{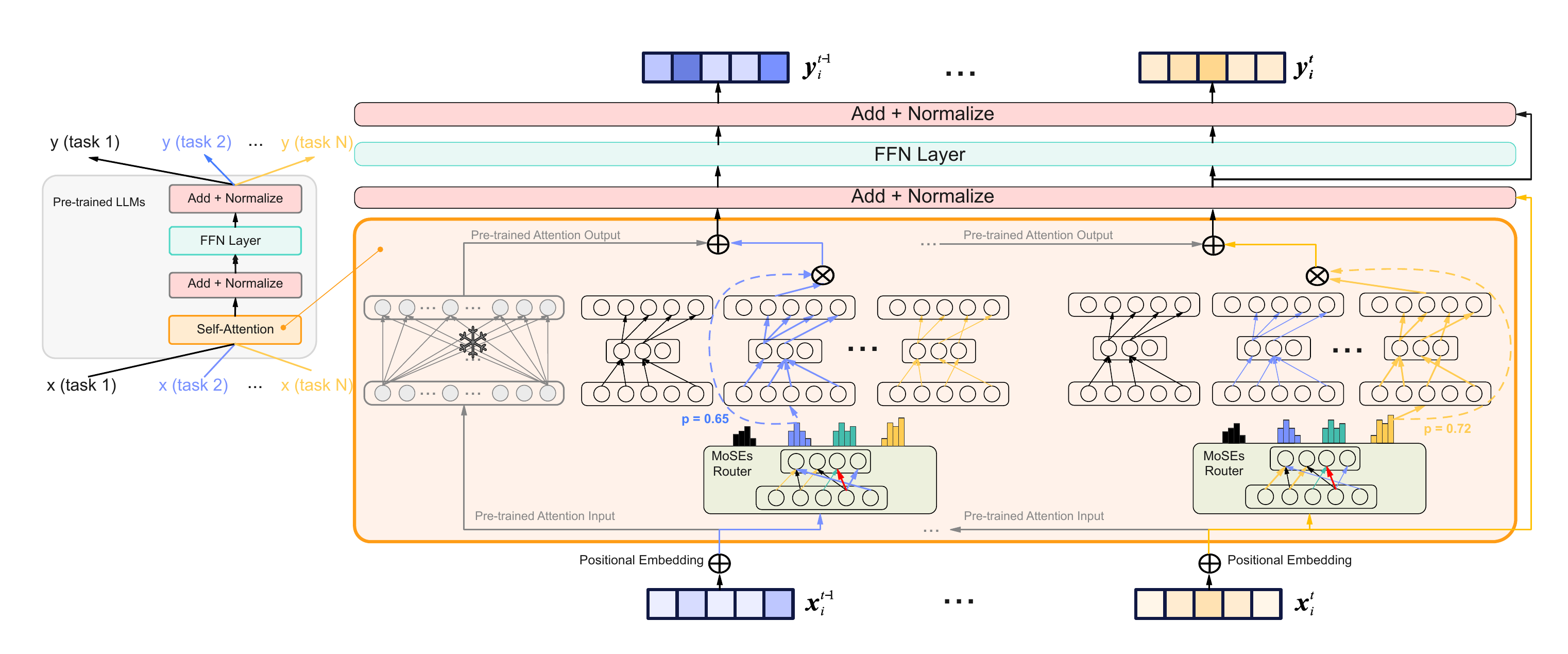}
    \caption{\small \textbf{Mixtures of SubExperts (MoSEs)}: The Self-Attention layer is fine-tuned by MoSEs to operate on task-specific tokens. Given, $\bm{x}_i^{t-1}$ and $\bm{x}_i^{t}$, the MoSEs router adaptively distributes them across $N$ subexperts. Each token is routed to the most relevant subexperts, and the final output is computed as the weighted sum of the selected subexpert outputs, where the weights correspond to the router gate values (e.g., $p = 0.65$ or $p = 0.72$).}
    \label{fig:arch_moses}
    \vspace{-10pt}
\end{figure*}

To address the issues led by the classical LoRA and MoE, we propose a new continual learning framework, \textit{Mixtures of SubExperts (MoSEs)}, designed for large-scale continual language models, as shown in \Cref{fig:arch_moses}. MoSEs enable scalable and efficient continual learning by combining sparse mixture-of-subexperts' representations, task-specific sparse routing, and parameter-efficient fine-tuning at attention layers. To control model capacity growth, MoSEs select sparse subexperts and adaptively reuse previously learned weights. Unlike traditional methods that require loading separate weights per task, our MoSEs simplify task management through binary subnets and prompt-based control. To avoid routing instability and expert collapse, MoSEs apply top-c\% subexpert sparsity selection without freezing previously learned parameters, effectively preserving prior knowledge while enabling robust and stable task adaptation.

\subsection{Mixtures of SubExperts}
Similar to LoRA, sparse MoSEs are augmented with pre-trained parameters at selected transformer attention layers. Each MoSE attention layer consists of a pool of $N$ sub-neural experts and a trainable \textbf{S}parse \textbf{R}outing (\textbf{SR}) function. For a given input representation $\bm{x}$, the sparse router computes a score vector $\hat{R}(\bm{x}) \in \mathbb{R}^N$ that weights the relevance of each \textbf{S}ub\textbf{E}xpert (\textbf{S}\textbf{E}) $\hat{E}(\bm{x})$. A sparse top-$k$ selection is applied to activate only the most relevant experts:
\begin{equation}
\text{MoSE}_{t}^{l}(\bm{x}) = \sum_{j \in \text{TopK}(\hat{\bm{R}}^l(\bm{x}))} \hat{R}^l_{\tilde{\bm{\theta}} \odot \bm{\delta}^t_j}(\bm{x}) \cdot \hat{E}^l_{\tilde{\bm{\theta}} \odot \bm{\xi}^t_j}(\bm{x}),
\label{eq:routing}
\end{equation}
where $\hat{R}^l_{\tilde{\bm{\theta}} \odot \bm{\delta}^t_j}(\bm{x})$ is the \textbf{SR} weight represented by sparse parameters $\tilde{\bm{\theta}} \odot \bm{\delta}^t_j$ and $\hat{E}_{\tilde{\bm{\theta}} \odot \bm{\xi}^t_j}$ is the $j$-th \textbf{S}\textbf{E} parameterized by $\tilde{\bm{\theta}} \odot \bm{\xi}^t_j$ at $t$-th task; binary masks $\bm{\delta}^t$ and $\bm{\xi}^t$ for the SR and SE are constructed via top-$c\%$ selection over their corresponding weight score functions $\bm{s}_{\bm{\delta}^t}$ and $\bm{s}_{\bm{\xi}^t}$, respectively. The underlying parameters $\tilde{\bm{\theta}}$ are shared and learnable. Only selected subexperts are involved in the forward and backward passes, ensuring computational efficiency and localized updates.

To enable task-adaptive modeling, sparse routing assigns each task a distinct subset of sparse experts. Upon arrival of a new task $t$, we assign a layer-wise task-specific expert subset $\text{MoSE}^l_{\tilde{\bm{\theta}} \odot \bm{m}^t_j}$, train only the sparse parameters of $\hat{E}_{\tilde{\bm{\theta}} \odot \bm{\xi}^t_j}$ and $\hat{R}_{\tilde{\bm{\theta}} \odot \bm{\delta}^t_j}$. The hidden state $\bm{h}^l$ at the $l$-th layer is adapted by MoSE, $\bm{h}^l = \bm{x} \bm{\theta}^l + \beta \cdot \text{MoSE}^l_{t}(\bm{x})$, where the pre-trained weight $\bm{\theta}^l \in \mathbb{R}^{m \times n}$ and mixtures of subexperts, MoSE$^l$ at layer $l$; $\beta = \alpha / r$ is determined by the adaptation rate $\alpha$ and rank size $r$. This approach ensures that each task operates within a minimally interfering subnetwork, enabling task-specific adaptation without overwriting prior knowledge $\bm{\theta}^l$. Task interference is further reduced by enforcing orthogonality in sparse expert usage patterns via the following task prompts.

\subsection{Task Adaptation of SubExperts}
The task-adaptive hidden representation $\tilde{\bm{h}}^l$ in the $l$-th attention layer is further governed by a set of learnable task-specific keys $\bm{K} = \{\bm{k}_t\}_{t=1}^{\mathcal{T}}$, where each $\bm{k}_t \in \mathbb{R}^D$ is designed to capture the representative features of task $t$ among the total $\mathcal{T}$ tasks. Unlike conventional methods that concatenate prompt parameters into the input stream, our approach maintains the original sequence length of the input hidden state $\bm{x}^l$ to ensure computational efficiency. Specifically, the task-adaptive representation is obtained by leveraging the task key $\bm{k}_t$ to guide the Mixture of SubExperts (MoSEs) via a conditioned routing mechanism, formulated as follows:
\begin{equation}
\tilde{\bm{h}}^l = \bm{x}^l \bm{\theta}^l + \beta \cdot \text{MoSE}^l(\bm{x}^l; \bm{k}_t)
\label{eq:task_adaptation}
\end{equation}
where $\text{MoSE}^l(\cdot)$ utilizes $\bm{k}_t$ to dynamically select and combine specialized parameters. This allows the MoSEs to facilitate efficient knowledge transfer and retention without the memory and computational overhead of additional prompt tokens.

\begin{wraptable}{r}{0.53\textwidth} 
    \centering
    \vspace{-15pt} 
    \begin{minipage}{0.52\textwidth} 
        \begin{algorithm}[H] 
            \caption{MoSEs at training time}
            \label{alg:algo_moses_train}
            \small
        \begin{algorithmic}[1]
            \STATE \textbf{Input}: MoSEs $f_{\bm{\theta} \odot {\bm{m}}}$, where $\bm{m} = \{\bm{\delta}, \bm{\xi} \}$
            \STATE ~ Training set $\{\{\bm{x}_{i,t}, y_{i,t}\}_{i=1}^{n_t} \}_{t=1}^{\mathcal{T}}$ 
            \STATE ~ Task keys $\bm{K}=\{\bm{k}_t\}_{t=1}^{\mathcal{T}}$, Masks $\bm{M} = \{\bm{m}_t\}_{t=1}^{\mathcal{T}}$
            \STATE ~ Learnable $\tilde{\bm{\theta}}$, Scores $\bm{s}$, Epochs $\mathcal{E}_t$
            \STATE \textbf{Initialize}: $\bm{\theta}, {\bm{M}}, \bm{K}$
            \FOR{task $t = 1, \cdots , \mathcal{T}$}
                \STATE Select task-specific key $\bm{k}_t$. 
                \STATE Config $f_{\tilde{\bm{\theta}} \odot \bm{m}_t}$ at [$start_e, end_e$] using $\bm{k}_t$. 
                \FOR{epoch $e=1, \cdots, \mathcal{E}_t$} 
                    \STATE Draw mini-batch $B = \{(\bm{x}_{i, t}, y_{i,t})\}_{i=1}^{n_t}$
                    \FOR{($\bm{x}, \bm{y}$) in $B$} 
                        \STATE Get binary masks $\bm{m}_t$ from top-$c$\% scores $\bm{s}$.
                        \STATE $\mathcal{L}_{total} = \mathcal{L}_{task} + \lambda_{1} \mathcal{L}_{cont} + \lambda_{2} \mathcal{L}_{ortho}$ \; (\ref{eq:total_loss})
                    \ENDFOR
                    \STATE Update $\bm{M}, \bm{K}, \tilde{\bm{\theta}}, \bm{s}$ via back-prop.
                \ENDFOR
            \ENDFOR
        \end{algorithmic}
\end{algorithm}
    \end{minipage}
    \vspace{-20pt} 
\end{wraptable}

\subsection{Optimizations of MoSEs}
To ensure that the selected prompt keys remain semantically aligned while enhancing task discriminability, we use a \textbf{C}ontrastive \textbf{P}rompt \textbf{K}ey (\textbf{CPK}) Learning objective ($\mathcal{L}_{\text{cont}}$). This approach utilizes an InfoNCE-style loss that pulls the normalized input embedding $\hat{\bm{x}}_i$ toward its ground-truth prompt key $\hat{\bm{k}}_{t}$ while simultaneously pushing it away from all other keys $\{\hat{\bm{k}}_j\}_{j \neq t}$ in the pool. Formally, for a batch of size $B$ and a temperature parameter $\tau=0.1$, the loss is defined as $\mathcal{L}_{\text{cont}} = - \frac{1}{B} \sum_{i=1}^{B} \log \frac{\exp(\langle \hat{\bm{x}}_i, \hat{\bm{k}}_{t} \rangle / \tau)}{\sum_{j=1}^{N} \exp(\langle \hat{\bm{x}}_i, \hat{\bm{k}}_j \rangle / \tau)}$. This formulation forces the model to learn a sharper decision boundary, significantly reducing task-confusion errors during inference by penalizing high similarity with incorrect prompt keys. To further safeguard the structural integrity of the prompt pool, we introduce a Key Orthogonality Constraint ($\mathcal{L}_{\text{ortho}}$) defined as $\left\| \bm{K} \bm{K}^\top - \bm{I} \right\|_F^2$, where $\bm{K}$ is the prompt key matrix and $\bm{I}$ is the identity matrix. This term and the sparse routing (SR) weight $\hat{R}^l_{\tilde{\bm{\theta}} \odot \bm{\delta}^t_j}(\bm{x} \oplus \bm{k}_t)$ encourage each SubExpert's address to occupy a unique task direction in the embedding space, preventing representation collapse where multiple tasks cluster in the same region. The total training objective of MoSEs thus becomes as follows: 
\begin{equation}
\mathcal{L}_{\text{total}} = \mathcal{L}_{\text{task}} + \lambda_1 \cdot \mathcal{L}_{\text{cont}} + \lambda_2 \cdot \mathcal{L}_{\text{ortho}}, 
\label{eq:total_loss}
\end{equation}
where $\lambda_1$ and $\lambda_2$ balance the discriminative pull and geometric independence, shown in \Cref{alg:algo_moses_train}. During inference, \Cref{alg:algo_test}, the model selects the most correlated prompt key $t^* = \arg\max_{j} \langle \hat{\bm{x}}, \hat{\bm{k}}_j \rangle$, allowing MoSEs to route inputs through task-specific subnetworks dynamically without explicit task labels.

\section{Experiments}\label{sec:exp}
We validate our method on two large language benchmark datasets: TRACE and SuperNI, against CL baselines, demonstrating the novel parameter-efficient isolated continual finetuning method.

\subsection{Experimental Settings}
\paragraph{Datasets.} To evaluate the effectiveness of MoSEs in continual learning, we utilize the TRACE benchmark~\citep{wang2023trace}, a dataset specifically designed to be contamination-free, challenging for modern LLMs, and task-diverse. Within this framework, we incorporate a suite of traditional NLP tasks from SuperNI~\cite{wang2022super}, adhering to the evaluation protocol established by SEEKR~\cite{he2024seekr}. Specifically, SuperNI includes three datasets from each of the four primary task categories—Information Extraction, Question Answering, Summarization, and Sentiment Analysis. This configuration facilitates a rigorous assessment of knowledge retention and scalability across a diverse task stream while ensuring computational efficiency remains optimized.

\begin{table}[ht]
\centering
\vspace{-8pt}
\caption{\small \textbf{(TAG-CLL) Experimental results} on the TRACE benchmark with LLM backbones and parameters. $\ast$ denotes our reproduced results.}
\label{tab:trace_results}
\resizebox{\textwidth}{!}{%
\begin{tabular}{lccccccccc}
\toprule
\multirow{2}{*}{\textbf{Method}} & \multicolumn{3}{c}{\textit{mistralai / Mistral-7B-Instruct-v0.3}} & \multicolumn{3}{c}{\textit{meta-llama / LLaMA-2-7B-Chat}} & \multicolumn{3}{c}{\textit{google / Gemma-2B-it}} \\ \cmidrule(lr){2-4} \cmidrule(lr){5-7} \cmidrule(lr){8-10}
 & \textbf{ACC} $\uparrow$ & \textbf{BWT} $\uparrow$ & \textbf{\# Params} $\downarrow$ & \textbf{ACC} $\uparrow$ & \textbf{BWT} $\uparrow$ & \textbf{\# Params} $\downarrow$ & \textbf{ACC} $\uparrow$ & \textbf{BWT} $\uparrow$ & \textbf{\# Params} $\downarrow$ \\ \midrule

L2P$^\ast$ & $49.32_{\pm 0.8}$ & $-5.34_{\pm 0.6}$ & $1.97$M & $36.23_{\pm 0.8}$ & $-8.25_{\pm 0.8}$ & $1.97$M & $31.14_{\pm 1.2}$ & $-15.77_{\pm 0.7}$ & $0.50$M \\
DualPrompt$^\ast$ & $51.14_{\pm 1.2}$ & $-6.13_{\pm 0.5}$ & $2.01$M & $37.69_{\pm 1.2}$ & $-8.03_{\pm 0.8}$ & $2.01$M & $32.42_{\pm 1.0}$ & $-14.25_{\pm 0.5}$ & $0.54$M \\ \midrule
SeqLoRA & $46.94_{\pm 1.2}$ & $-11.41_{\pm 0.6}$ & $3.41$M & $34.30_{\pm 1.2}$ & $-18.50_{\pm 0.8}$ & $4.19$M & $31.89_{\pm 0.8}$ & $-15.28_{\pm 0.4}$ & $0.92$M \\

LoRA$^{\ast}$ & $50.62_{\pm 0.9}$ & $-9.61_{\pm 0.6}$ & $3.41$M & $40.32_{\pm 0.9}$ & $-8.58_{\pm 0.5}$ & $4.19$M & $32.44_{\pm 0.9}$ & $-14.36_{\pm 0.6}$ & $0.92$M \\

HiDeLoRA & $51.81_{\pm 0.9}$ & $-6.25_{\pm 0.3}$ & $3.41$M & $41.60_{\pm 0.8}$ & $-7.12_{\pm 0.4}$ & $4.19$M & $33.25_{\pm 0.9}$ & $-13.66_{\pm 0.5}$ & $0.92$M \\
O-LoRA & $52.02_{\pm 0.8}$ & $-8.13_{\pm 0.6}$ & $3.41$M & $42.78_{\pm 0.8}$ & $-7.16_{\pm 0.4}$ & $4.19$M & $33.73_{\pm 0.8}$ & $-12.36_{\pm 0.4}$ & $0.92$M \\
TreeLoRA & $54.77_{\pm 1.1}$ & $-3.77_{\pm 0.4}$ & $3.41$M & $43.52_{\pm 1.0}$ & $-3.46_{\pm 0.4}$ & $4.19$M & $33.41_{\pm 0.9}$ & $-8.50_{\pm 0.5}$ & $0.92$M \\ 
GainLoRA(InfLoRA)$^{\ast}$ & $54.83_{\pm 1.0}$ & $-3.63_{\pm 0.4}$ & $2.12$M & $49.01_{\pm 0.9}$ & $-3.35_{\pm 0.4}$ & $2.51$M & $33.53_{\pm 0.8}$ & $-8.48_{\pm 0.4}$ & $0.66$M \\
\midrule

\rowcolor[HTML]{EFEFEF}
\textbf{MoSEs} (ours) & $\mathbf{54.97_{\pm 0.7}}$ & \textbf{$-1.83_{\pm 0.4}$} & $\mathbf{1.45}$M & $\mathbf{52.76_{\pm 0.5}}$ & $\mathbf{-0.50_{\pm 0.1}}$ & $\mathbf{2.15}$M & $\mathbf{35.93_{\pm 0.2}}$ & $\mathbf{-5.02_{\pm 0.1}}$ & $\mathbf{0.42}$M \\
\bottomrule
\end{tabular}%
}
\vspace{-8pt}
\end{table}

\noindent
\textbf{Baselines.} The performance of MoSEs is evaluated particularly against LoRA-family baselines on the two TRACE and SuperNI benchmark datasets. These comparison approaches include the following methods: (1) prompt-based methods, L2P~\cite{wang2022learning} and DualPrompt\cite{wang2022dualprompt}, which maintain a bank of prompts chosen; (2) LoRA-based Sequential Fine-Tuning (SeqLoRA)~\cite{hu2021lora}, LoRA~\cite{hu2021lora}, HiDeLoRA~\cite{wang2025hide}, O-LoRA~\cite{wang2023olora}, InfLoRA~\cite{liang2024inflora}, GainLoRA~\cite{liang2025gated}, and TreeLoRA~\cite{qian2025treelora}. We set the LoRA-family's hyperparameters as $r=8, \alpha=32$ (4.19M parameters). We also include conventional replay-based methods such as Replay, DER++~\cite{buzzega2020dark}, and SEEKR~\cite{he2024seekr} to show the effectiveness of MoSEs, particularly on the SuperNI benchmark dataset (the Appendix of \Cref{tab:app_task_sequence}).

\begin{wraptable}{r}{0.53\textwidth}
\centering
\vspace{-8pt} 
\caption{\small \textbf{(TAG-CLL) Comparison with baselines on the SuperNI benchmark} using LLaMA-2-7B. \textbf{MoSEs} W/O[0-1], E2T2 ($r=2,\alpha=8$). $\ast$ denotes our reproduced results.}
\label{tab:superni_comparison}
\setlength{\tabcolsep}{1pt} 
\resizebox{0.50\textwidth}{!}{
\begin{tabular}{lcccccc}
\toprule
 \multirow{2}{*}{\textbf{Method}} &  &  & \multicolumn{2}{c}{\textbf{Order1}} & \multicolumn{2}{c}{\textbf{Order2}} \\ 
 \cmidrule(lr){2-4}\cmidrule(lr){4-7}
 & \multicolumn{1}{c}{\textbf{Replay Buffer}}  &  \textbf{\# Params} $\downarrow$ & \textbf{ACC} $\uparrow$  & \textbf{BWT} $\uparrow$    & \textbf{ACC} $\uparrow$  & \textbf{BWT} $\uparrow$  \\
 \midrule
L2P$^\ast$                & - & 2.95M & 32.71 & -22.34    & 31.00 & -23.82 \\
DualPrompt$^\ast$  & - & 3.01M  & 26.47 & -20.00    & 25.96 & -21.40 \\
\midrule
SeqLoRA$^\ast$             & - & 6.29M & 42.12 & -19.71    & 42.26 & -19.78 \\ 
LoRA$^\ast$                & - & 4.19M & 43.12 & -18.54    & 43.27 & -18.62 \\ 
HiDeLoRA$^\ast$            & - & 4.19M & 44.67 & -18.33    & 44.56 & -18.46 \\ 
O-LoRA$^\ast$              & - & 4.19M & 45.12 & -17.53    & 45.30 & -17.62 \\ 
Tree-LoRA$^\ast$           & - & 4.19M & 46.54 & -16.21    & 46.85 & -16.88 \\ 
GainLoRA(InfLoRA)$^\ast$   & - & 2.51M & 51.21 & -11.32    & 50.23 & -11.73 \\
\midrule 
\rowcolor[HTML]{EFEFEF}
\textbf{MoSEs} (ours), c=29.0\% & -    & \textbf{2.25}M & \textbf{60.65} & ~\textbf{+2.40} & \textbf{59.81} & ~\textbf{+1.34} \\ \midrule
Replay                         & 1\%          & 6.46B    & 55.00 & -4.27     & 54.78 & (-5.31) \\
DER++                          & 1\%          & 6.46B    & 55.89 & -4.51     & 53.48 & (-5.01) \\
SEEKR                          & 1\%          & 6.46B    & 57.04 & -3.15     & 58.26 & (-2.52) \\ \midrule
MT                             & -            & 6.46B    & \multicolumn{4}{c}{61.27} \\ 
MT of MoSEs                    & -            & 2.36M    & \multicolumn{4}{c}{61.89} \\ \bottomrule
\end{tabular}}
\vspace{-10pt} 
\end{wraptable}

\noindent
\textbf{Training \& Inference.}
Our MoSEs framework's performance and behavior are thoroughly evaluated and analyzed across two distinct continual learning settings under the same experimental environments of TreeLoRA~\cite{qian2025treelora}, i.e., learning rate: Task-Agnostic Continual Language Learning (TAG-CLL) and Task-Aware Continual Language Learning (TA-CLL), where the model is explicitly provided with the specific task ID during both training and inference. TA-CLL settings remain highly relevant in real-world applications where clear task boundaries exist, such as domain-specific expert systems. Conversely, TAG-CLL presents a more challenging and realistic setting: although the model receives sequential task IDs during training, it must infer the task autonomously during inference by adapting a task ID based on the input. This gap between training and inference in TAG-CLL is bridged by leveraging task-relevant prompts, allowing the model to implicitly recognize the task from the input data without an explicit task ID during testing. To demonstrate the optimality of our framework, we evaluate MoSEs using both Cosine Similarity (CS) and our proposed Contrastive Prompt Key (CPK). The training details are shown in the Appendix of \Cref{tab:app_moses_setting}.

\subsection{Performances}
\noindent
\textbf{Performances.} 
MoSEs (ours) establishes a new state-of-the-art across the TRACE and SuperNI benchmarks by simultaneously maximizing accuracy and minimizing forgetting while utilizing a fraction of the parameters required by baseline methods. On the TRACE benchmark, as stated in \Cref{tab:trace_results}, it achieves a breakthrough in knowledge retention across various backbones, reaching a Backward Transfer (BWT) as high as $-0.50$, significantly outperforming the $-3.35$ of the strongest baseline, GainLoRA. This superiority is further amplified on the SuperNI benchmark, as shown in \Cref{tab:superni_comparison}, where MoSEs is the only method to achieve positive Backward Transfer (+2.40), indicating a paradigm shift from merely avoiding forgetting to enabling positive knowledge transfer. Interestingly, while TRACE consists of heterogeneous datasets requiring tighter knowledge isolation, the SuperNI benchmark is composed of more similar tasks; by utilizing slightly more parameters (2.25M vs 2.15M) to facilitate broader parameter sharing in this task-similar environment, MoSEs successfully unlocked the ability for new tasks to actually enhance previous ones. Remarkably, MoSEs delivers this performance, nearly reaching the Multi-Task (MT) upper bound, while effectively outperforming replay-based methods that rely on massive 6.46B parameter backbones and external buffers. This unique combination of high accuracy and extreme efficiency demonstrates that the sparse sub-routing architecture successfully bridges the performance gap between sequential continual learning and simultaneous joint training.

\begin{figure}[ht]
    \centering
    \small
    \vspace{-10pt}
    \setlength{\tabcolsep}{0pt}{%
    \begin{tabular}{cc}

    \includegraphics[width=0.47\columnwidth]{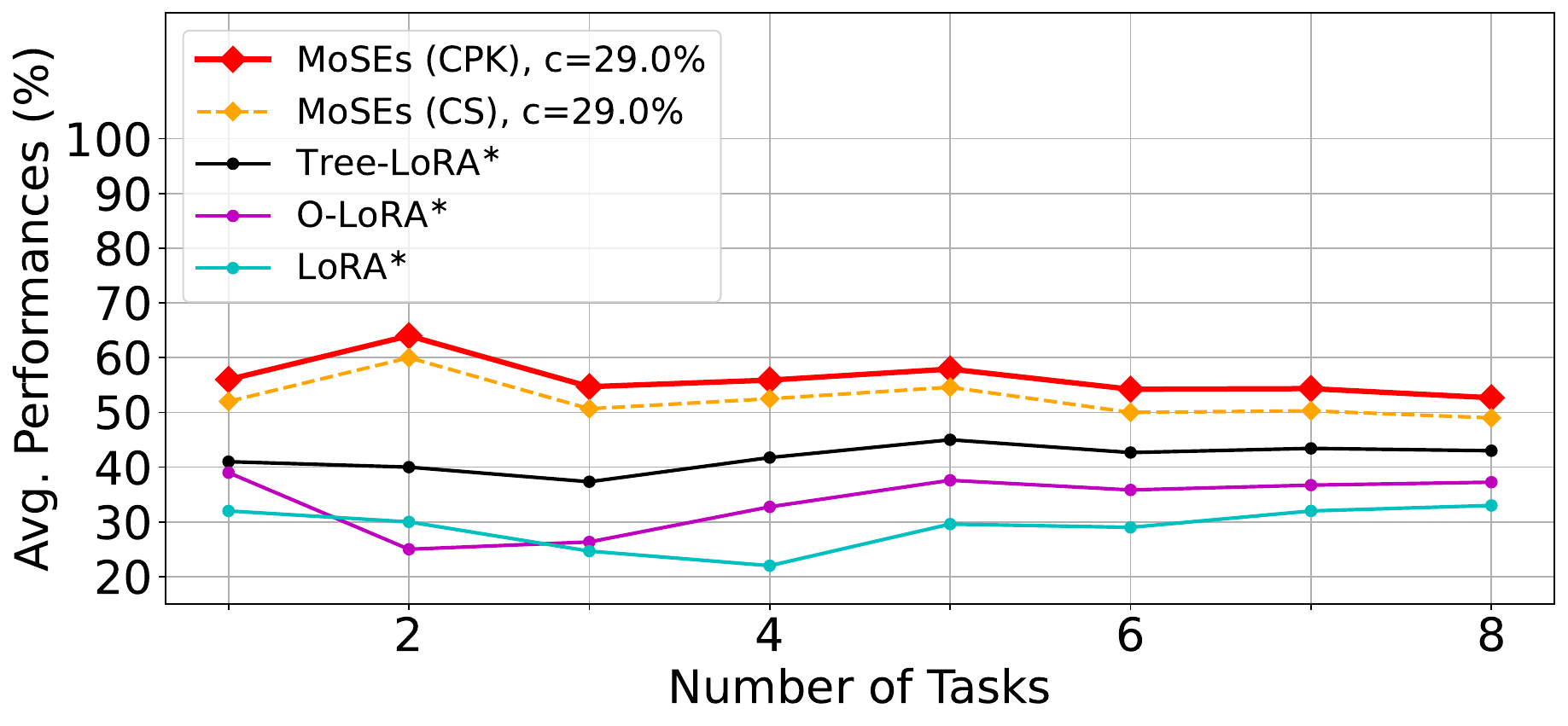} &     
    \includegraphics[width=0.45\columnwidth]{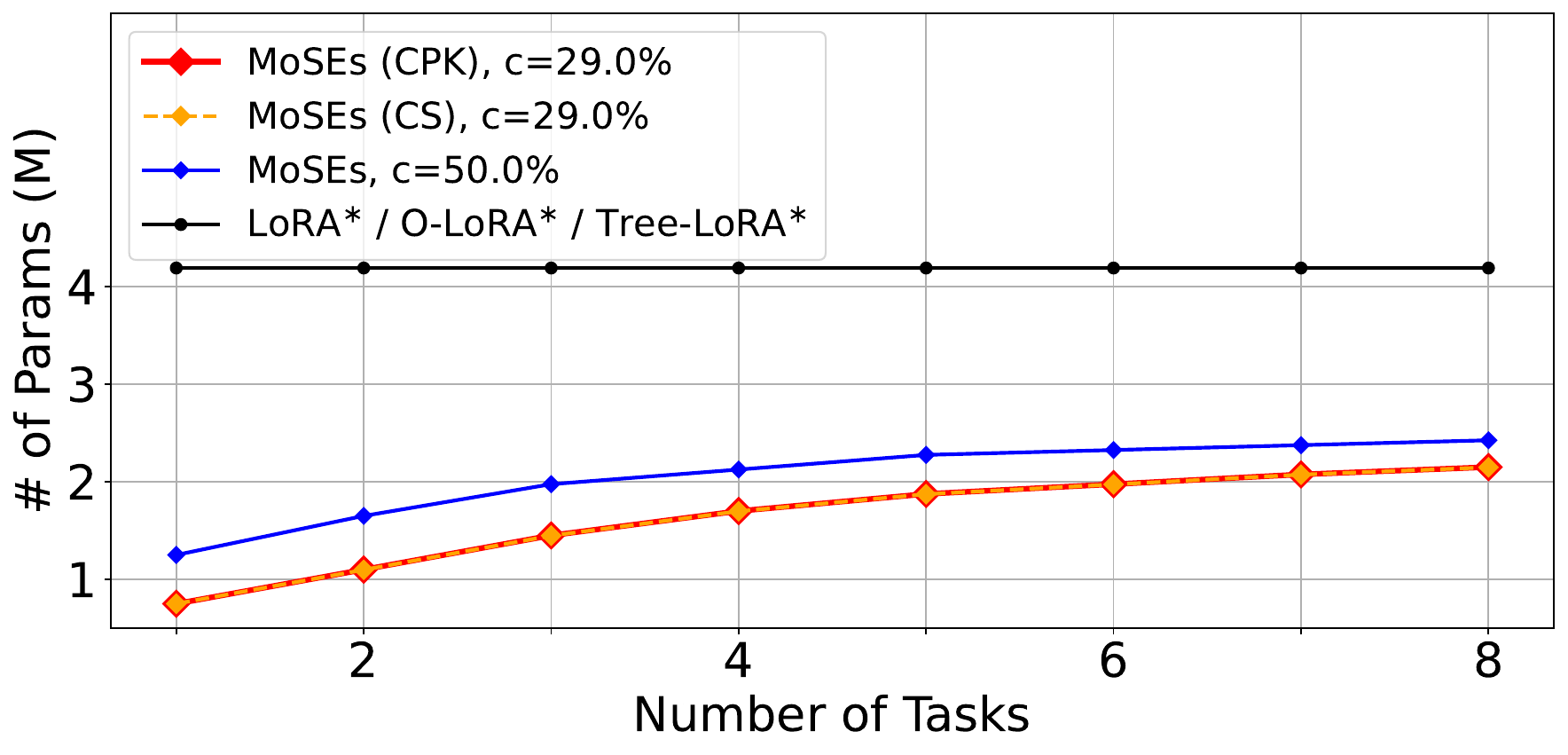} \\
    \small (a) Average Performances of MoSEs & \small (b) Cumulative Training Parameters (Million, M). \\
    \end{tabular}
    }
    \caption{\small \textbf{(TAG-CLL) Average Performances and Model Capasity of MoSEs} on the TRACE.}
    \label{fig:perf_capasity}
    \vspace{-0.15in}
\end{figure}

\noindent
\textbf{Parameter-Efficient CL.}
The results presented in \Cref{fig:perf_capasity} highlight the strong performance and parameter efficiency of MoSEs across increasing numbers of tasks in the TRACE benchmark. As shown in \Cref{fig:perf_capasity}(a), MoSEs (CPK) with a 29.0\% expert selection ratio consistently outperform other baselines - such as LoRA, O-LoRA, and MoE - in terms of average performance, maintaining higher accuracy as the number of tasks grows. Meanwhile, \Cref{fig:perf_capasity}(b) demonstrates that MoSEs achieve this superior performance with significantly fewer trainable parameters. Specifically, MoSEs with $c=29.0\%$ require notably fewer parameters and scale more efficiently compared to other baselines, even as tasks accumulate. This demonstrates MoSEs’ ability to strike a favorable balance between effectiveness and computational efficiency.

\subsection{Ablation Studies on Minimal Task-Interference}

\noindent
\textbf{Sparsity Analysis.}
\Cref{tab:til_sparse_mose} highlights the effectiveness of sparsity in MoSEs under the task-aware continual language learning (TA-CLL) setting on the TRACE. Interestingly, the sparse MoSEs configuration with c=30.0\% achieves the highest average performance of 54.9\%, outperforming denser variants with c=40.0\% and c=50.0\%, which yield 53.6\% and 54.7\%, respectively. In addition to superior accuracy, the c=30.0\% model also exhibits favorable backward transfer (BWT) of +0.05\%, while maintaining the smallest parameter (2.17M). Despite slight gains in BWT at higher sparsity (e.g., +0.16\% for c=50.0\%), the performance does not improve accordingly. These findings demonstrate that properly sparse parameters of previously learned and newly learned parameters in the attention layer (e.g., c=30.0\%) strike an optimal balance between accuracy, stability, and efficiency in CL.

\begin{table}[ht] 
    \centering
    \vspace{-8pt}
    \begin{minipage}{0.48\textwidth}
    \centering
    \caption{\small \textbf{(TA-CLL) Sparse MoSEs} (W/O[0-1], E2T2 ($r=2,\alpha=8$)) on the TRACE.}
    \label{tab:til_sparse_mose}
    \resizebox{0.99\textwidth}{!}{
    \begin{tabular}{lcccc}
    \toprule
    \textbf{Method} &  \textbf{ACC}  & \textbf{BWT} & \textbf{\# Params} & \textbf{Train / Test} \\
    \midrule
    \rowcolor[HTML]{EFEFEF}
    \textbf{MoSEs}, \textbf{c=30.0\%} & \textbf{54.9} & \textbf{+0.05 \%} & \textbf{2.17M}  & \textbf{20.16h} / \textbf{1.20h} \\
    \textbf{MoSEs}, c=40.0\% &  {53.6}  & -0.71 \% & 2.35M & 20.17h / 1.20h \\
    \textbf{MoSEs}, c=50.0\% &  {54.7}  & +0.16\% & 2.46M & 20.18h / 1.20h \\
    \bottomrule
    \end{tabular}
    }
    \end{minipage}
    \hfill 
    \begin{minipage}{0.48\textwidth}
    \centering
    \caption{\small \textbf{(TA-CLL) Layer-wise MoSEs} c=30.0\%, E2T2 ($r=2, \alpha=8$) on the TRACE. W[0\text{–}31]: all-layer tuning; W/O: layer skipping.}
    \label{tab:layerwise_results}
    \resizebox{0.99\textwidth}{!}{
    \begin{tabular}{lcccc}
    \toprule
    \textbf{Method} &  \textbf{ACC} & \textbf{BWT} & \textbf{\# Params} & \textbf{Train / Test} \\
    \midrule
    
    \textbf{MoSEs}, W[0-31]  & 52.6 & -1.50 \% & 2.33M & 20.45h / 1.21h \\
    \rowcolor[HTML]{EFEFEF}
    \textbf{MoSEs}, W/O[0-1] & \textbf{54.9}  & \textbf{+0.05} \% & 2.17M & 20.20h / 1.20h \\
    \textbf{MoSEs}, W/O[0-2] & 54.3  & -1.30 \% & 2.11M & 20.16h / 1.19h \\
    \bottomrule
    \end{tabular}
    }
    \end{minipage}
    \vspace{-10pt}
\end{table}

\noindent
\textbf{Layer-wise Efficiency.}
\Cref{tab:layerwise_results} presents the layer-wise performance of MoSEs ($c=30\%$) and reveals that selective subnet tuning contributes to efficient performance with minimal parameter overhead. The full-layer version, {MoSEs W[0-31]}, achieves an average accuracy of 52.6 with a BWT of {-1.50\%} and parameter count of 2.33M. Interestingly, MoSEs W/O[0-1] improves both average performance ({54.9}) and BWT ({+0.05\%}) while reducing parameters to {2.17M}, indicating that early layers may not be essential for task adaptation and can even lead to forgetting. As more lower layers are excluded, e.g., W/O[0-2], the model size decreases further to 2.11M, but at the cost of degraded average accuracy (54.3) and worsened forgetting (BWT up to {-1.30\%}). This trend demonstrates a clear trade-off: while excluding lower layers can improve efficiency, overly aggressive layer skipping degrades overall knowledge retention and transfer. These findings confirm that MoSEs are robust and efficient, protecting universal language understanding of transformers, especially when low-level layers are strategically frozen. The optimal balance is achieved by MoSEs W/O[0-1], which combine strong performance, positive transfer, and parameter savings.

\begin{table}[ht] 
    \centering
    \vspace{-7pt}
    \begin{minipage}{0.46\textwidth}
    \centering
    \caption{\small \textbf{(TA-CLL) Expert Performances } of MoSEs, c= 30.0\% ($r=2, \alpha=8$) on the TRACE.}
    \label{tab:trace_experts}
    \resizebox{0.98\textwidth}{!}{
    \begin{tabular}{lcccc}
    \toprule
    \textbf{Method} & \textbf{ACC} & \textbf{BWT} & \textbf{\# Params} & \textbf{Train / Test} \\
    \midrule
    
    \rowcolor[HTML]{EFEFEF}
    \textbf{MoSEs}, \textbf{E2T2} & \textbf{54.9}  & \textbf{+0.05 \%} & \textbf{2.17}M  & \textbf{20.20h} / \textbf{1.20h} \\
    
    \textbf{MoSEs}, E3T2 & 50.2  & -2.00 \% & 2.45M & 20.20h / 1.20h \\
    
    \textbf{MoSEs}, E3T3 & 50.4  & -2.70 \% & 2.45M & 20.20h / 1.20h \\
    
    
    
    \bottomrule
    \end{tabular}}
    \vspace{-0.01in}
    \end{minipage}
    \hfill 
    \begin{minipage}{0.52\textwidth}
    \centering
    \caption{\small \textbf{(TA-CLL) Rank-wise MoSEs}, c=30.0\%, E2T2 on the TRACE. $r$: low-rank size; $\alpha$: a scaling factor.}
    \label{tab:trace_rank}
    \resizebox{0.98\textwidth}{!}{
    \begin{tabular}{lccccccccccccc}
    \toprule
    \textbf{Method} &\textbf{ACC} & \textbf{BWT} & \textbf{\# Params} & \textbf{Train / Test} \\
    \midrule
    \rowcolor[HTML]{EFEFEF}
    \textbf{MoSEs}, $\bm{r=2},\bm{\alpha=8}$ &  \textbf{54.9}  & \textbf{+0.05 \%} & \textbf{2.17}M  & \textbf{20.20h} / \textbf{1.20h} \\
    
    \textbf{MoSEs}, $r=3,\alpha=12$ &  52.8  & -0.99 \% & 3.45M  & 20.20h / 1.20h \\
    
    \textbf{MoSEs}, $r=4,\alpha=16$ &  50.8  & -1.22 \% & 4.38M  & 20.21h / 1.20h \\

    
    
    \bottomrule
    \end{tabular}}
    \end{minipage}
    \vspace{-8pt}
\end{table}

\noindent
\textbf{Effect of Expert Configuration.}
\Cref{tab:trace_experts} illustrates how varying the number of experts ($E\#$) and top-K selections ($T\#$) affects the performance and efficiency of MoSEs. The baseline setting \textbf{MoSEs, E2T2}, with two experts and top-2 selection, achieves the best average performance ({54.9}) and positive backward transfer ({+0.05\%}) with only {2.17M} parameters and a test time of {1.20h}, indicating an optimal trade-off between capacity and knowledge retention. As the number of experts increases to three ({E3T2}, {E3T3}), both the parameter count ({2.45M}) and forgetting grow significantly (BWT drops to as low as {-2.70\%}), while the average accuracy declines (down to {50.2}). These results suggest that increasing expert diversity without sufficient selectivity can dilute task-specific knowledge and introduce interference. Overall, the configuration {E2T2} demonstrates the best balance across all metrics, affirming that a minimal yet well-structured mixture of experts is most effective.

\noindent
\textbf{Low-Rank and Scaling Factor.}
\Cref{tab:trace_rank} indicates that MoSEs with $r=2, \alpha=8$ achieve the best trade-off between performance and efficiency. It records the highest average score (54.9), maintains a \emph{positive backward transfer} (+0.05\%), and uses the least number of parameters (2.17M). Increasing the rank and scaling factor leads to larger models but \emph{worsens BWT} and typically results in \emph{lower performance}, suggesting diminishing returns with higher capacity.

\begin{figure}[ht]
    \centering
    \small
    \vspace{-10pt}
    \setlength{\tabcolsep}{0pt}{%
    \begin{tabular}{cc}

    \includegraphics[width=0.46\columnwidth]{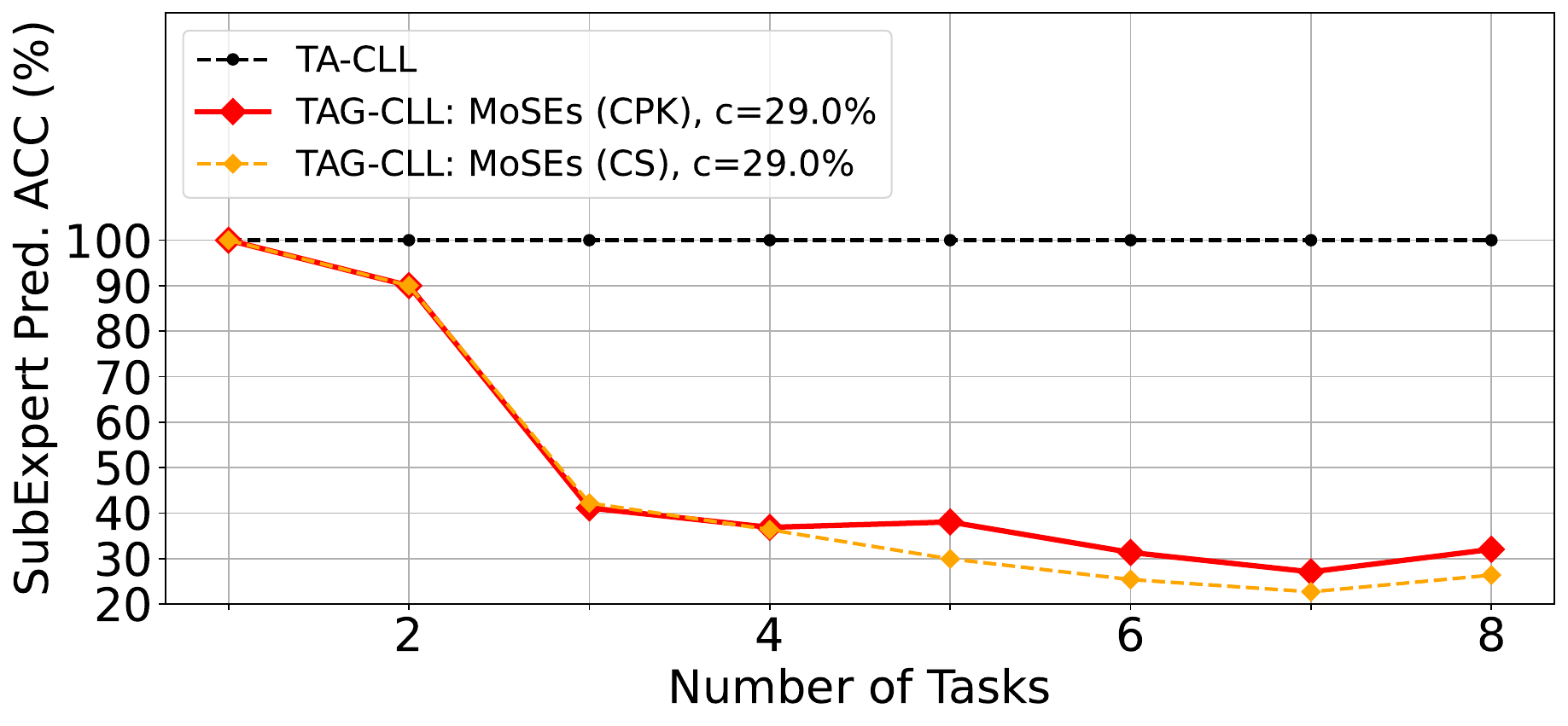} &     \includegraphics[width=0.50\columnwidth]{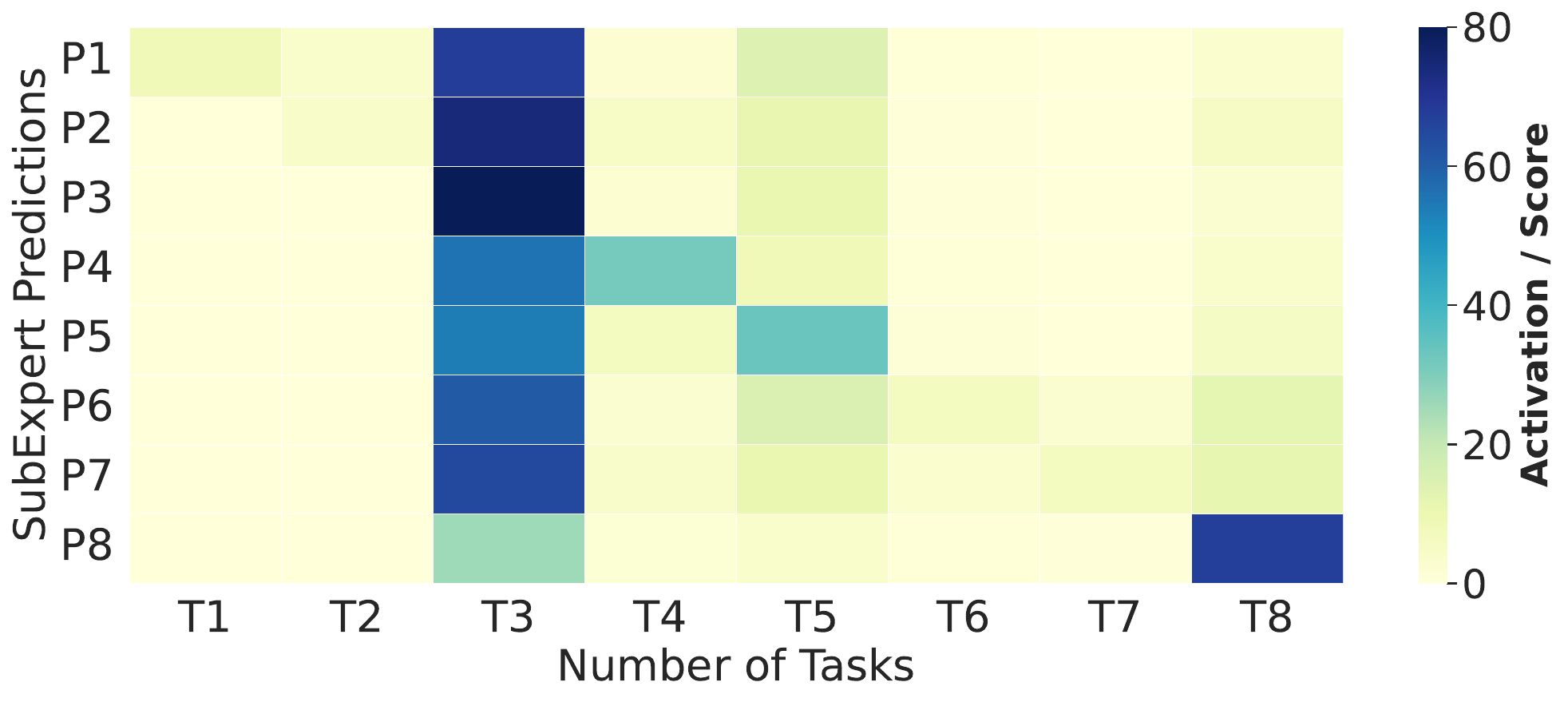} \\
    \small (a) SubExpert's ID Prediction Accuracy & \small (b) MoSEs (CPK) Routing Patterns. \\
    \end{tabular}
    }
    \caption{\small \textbf{SubExpert Predictions} \& \textbf{Routing Patterns} on the TRACE benchmark.}
    \label{fig:task_acc_trace}
    \vspace{-8pt}
\end{figure}

\noindent
\textbf{CPK Embedding.}
The robustness of MoSEs is exemplified by its sustained high performance (see \Cref{fig:perf_capasity}(a)) despite declining SubExpert's ID accuracy in \Cref{fig:task_acc_trace}(a). This resilience stems from a shared parameter architecture across SubExperts (see \Cref{fig:perf_capasity}(b)), creating a distributed knowledge manifold where related tasks leverage overlapping features even when primary expert assignments are imperfect. By utilizing CPK embeddings, the SubExpert's routing patterns in \Cref{fig:task_acc_trace}(b) maximize this utility through a dual-axis structural synergy: a vertical (column-wise) analysis shows each task dynamically distributing its 100\% activation capacity across a modular expert composition, while a horizontal (row-wise) analysis illustrates the specialized reusability of individual SubExperts. This mechanism allows MoSEs to bridge the gap between imperfect routing and optimal inference of TA-CLL, outperforming the CS and securing SOTA results across increasingly complex task streams.

\begin{wraptable}{r}{0.58\textwidth}
\centering
\vspace{-15pt} 
\caption{\small \textbf{(TAG-CLL) Component-wise Ablation Study} of MoSEs on the SuperNI benchmark (order1) in terms of Sub-Expert (\textbf{S}\textbf{E}), Sparse Routing (\textbf{SR}), and Pull Loss (\textbf{CPK}).}
\label{tab:component_ablation}
\resizebox{0.56\textwidth}{!}{
\begin{tabular}{ccc|cc|c}
\toprule
\textbf{SubExpert}       & \textbf{Sparse Routing} & \textbf{Pull Loss}            & \multicolumn{2}{c|}{\textbf{Performance}}        & \textbf{Efficiency} \\
(\textbf{S}\textbf{E}) & (\textbf{SR})           & (\textbf{CS} or \textbf{CPK}) & \textbf{ACC $\uparrow$} & \textbf{BWT $\uparrow$} & \textbf{\# Params} \\
\midrule
 -         & -          &  - &  37.68   &  +1.51\%   & 1.96M \\
\checkmark & -          &  - & 43.33 & +4.19\% & 1.89M \\
\checkmark & \checkmark & -  & 45.68 & +6.25\% & 2.25M \\

\checkmark & \checkmark & CS & 59.65 & +1.81\% & 2.25M \\
\rowcolor[HTML]{EFEFEF}
\checkmark & \checkmark & \textbf{CPK} & \textbf{60.65} & \textbf{+2.40\%} & \textbf{2.25M} \\
\bottomrule
\end{tabular}
}
\vspace{-8pt} 
\end{wraptable}

\noindent
\textbf{Ablation Study on MoSEs.}
MoSEs utilizes SubExperts (\textbf{S}\textbf{E} in \Cref{eq:routing}) to isolate task-specific knowledge, significantly boosting accuracy (from 37.68 to 43.33) while slightly reducing the initial parameters as shown in \Cref{tab:component_ablation}. The framework further enhances performance through Sparse Routing (\textbf{SR} in \Cref{eq:routing}) and specialized Pull Loss functions - specifically \textbf{CPK} represented in \Cref{eq:total_loss}, which yields a state-of-the-art performance of 60.65 on the SuperNI benchmark. By adaptively selecting and combining previously learned sparse parameters for new tasks, MoSEs facilitates effective knowledge transfer and ensures sublinear capacity growth, achieving superior efficiency and memory savings compared to baselines.



\section{Conclusion}
MoSEs (Mixtures of SubExperts) is a scalable continual learning framework that addresses the trade-off between catastrophic forgetting and linear parameter growth by integrating sparse sub-expert mixtures into transformer layers. By utilizing a task-specific sub-routing mechanism, MoSEs isolate knowledge within dedicated SubExperts to prevent parameter interference while adaptively combining previously learned sparse parameters to facilitate knowledge transfer. This architecture enables the model to achieve sublinear capacity growth and SOTA performance on the TRACE and SuperNI benchmarks, significantly improving knowledge retention and computational efficiency compared to standard PEFT methods.


\newpage
\bibliographystyle{plain}
\bibliography{reference}

\newpage
\appendix
\section{Appendix}

\subsection{Dataset Details}
To evaluate the effectiveness of MoSEs, we utilize the TRACE benchmark~\cite{wang2023trace}, which is characterized by its resistance to pre-training data contamination, its high difficulty for modern LLMs, and its broad task diversity. The task order is shown in \Cref{tab:app_task_sequence}. This evaluation is augmented by a suite of traditional NLP tasks from SuperNI~\cite{wang2022super}, following the SEEKR~\cite{he2024seekr} protocol. Specifically, SuperNI includes three datasets from each of four primary categories-Information Extraction, Question Answering, Summarization, and Sentiment Analysis—sampling $1,000$ training and $100$ testing instances per dataset. This comprehensive configuration ensures a rigorous assessment of knowledge retention and scalability across diverse task streams while maintaining optimal computational efficiency. The two task orders are stated in \Cref{tab:app_task_sequence}.

\begin{table}[ht]
\centering
\caption{Task sequence of different task orders.}
\label{tab:app_task_sequence}
\resizebox{0.85\textwidth}{!}{
\begin{tabular}{c|c|c}
\hline
\textbf{Order} & \textbf{Benchmark} & \textbf{Task Sequence} \\ \hline
 & TRACE benchmark & \begin{tabular}[c]{@{}c@{}}C-STANCE $\rightarrow$ FOMC $\rightarrow$ MeetingBank $\rightarrow$ Py150 $\rightarrow$ \\ ScienceQA $\rightarrow$ NumGLUE-cm $\rightarrow$ NumGLUE-ds $\rightarrow$ 20Minuten\end{tabular} \\ \hline
1 & SuperNI benchmark & \begin{tabular}[c]{@{}c@{}}task1572 $\rightarrow$ task363 $\rightarrow$ task1290 $\rightarrow$ task181 $\rightarrow$ task002 $\rightarrow$ task1510 $\rightarrow$ \\ task073 $\rightarrow$ task748 $\rightarrow$ task511 $\rightarrow$ task591 $\rightarrow$ task195 $\rightarrow$ task875\end{tabular} \\ \hline
2 & SuperNI benchmark & \begin{tabular}[c]{@{}c@{}}task748 $\rightarrow$ task073 $\rightarrow$ task1572 $\rightarrow$ task195 $\rightarrow$ task591 $\rightarrow$ task363 $\rightarrow$ \\ task1510 $\rightarrow$ task181 $\rightarrow$ task511 $\rightarrow$ task002 $\rightarrow$ task1290 $\rightarrow$ task875\end{tabular} \\ \hline
\end{tabular}}
\end{table}

\subsection{Training Details}

\noindent
\textbf{Baselines \& MoSEs.}
The continual learning performance of our proposed MoSEs method is evaluated against five baselines on the TRACE and SuperNI benchmark datasets. These comparison approaches include a non-training method, In-Context Learning (ICL)~\cite{brown2020language}, which uses a 6-shot prompt engineering setting with task demonstrations; and four training methods: Single Full-Parameter Fine-Tuning (Single FT), which trains all model parameters per task; LoRA-based Sequential Fine-Tuning (LoRA~\cite{hu2021lora}, O-Lora~\cite{wang2023olora}, HiDeLoRA~\cite{wang2025hide}, SAPT~\cite{zhao2024sapt}, TASL~\cite{feng2024tasl}, and TreeLoRA~\cite{qian2025treelora}), an efficient tuning technique that only fine-tunes low-rank matrices ($r=8, \alpha=32$, 4.19M parameters); Orthogonal Low-Rank Adaptation (O-LoRA)~\cite{wang2023olora}, which imposes orthogonality constraints on the update matrices to minimize task interference during sequential learning; and Mixture of Experts-based Sequential Fine-Tuning (MoE)~\cite{shazeer2017outrageously, lepikhin2020gshard, fedus2022switch}, which uses sparse routing to expert subnetworks to maintain fixed capacity without raising inference costs. Ultimately, our MoSEs demonstrates its effectiveness using the E2T2 configuration (2 experts, top-2 selection).

\begin{table}[ht]
\centering
\caption{\small \textbf{(TAG-CLL) Experimental settings} on the TRACE/SuperNI benchmark with LLM backbones.}
\label{tab:app_moses_setting}
\resizebox{\textwidth}{!}{%
\begin{tabular}{lccc}
\toprule
\textbf{MoSEs} (ours) & \textit{mistralai / Mistral-7B-Instruct-v0.3} & \textit{meta-llama / LLaMA-2-7B-Chat} & \textit{google / Gemma-2B-it} \\ 
Benchmark & TRACE & TRACE / SuperNI & TRACE \\
\midrule

Learning rate & 1e-4 & 1e-4 & 1e-4 \\ \midrule

Sparsity & c=15.0\% &  c=29.0\% & c=13.0\%  \\
Fine-tuning layers & W/O[0-1] & W/O[0-1] & W/O[0-1] \\
$\text{Rank}, \alpha$ & $r=3$, $\alpha=12$ & $r=2$, $\alpha=8$ & $r=3$, $\alpha=12$ \\ 
E\#T\# & E3T3 & E2T2 & E3T3\\   \midrule
Accumulated capacity & 50.0\% & 86.1\% / 90.0\% & 70.2 \% \\
\# of params & \textbf{1.45M} & \textbf{2.15M} / \textbf{2.25M}  & \textbf{0.42M} \\
\bottomrule
\end{tabular}%
}\label{tab:backbone_hyper}
\end{table}

\textbf{Training.} All MoSEs were trained with the same experimental settings~\cite{qian2025treelora}. The specific hyperparameters of MoSEs are given in \Cref{tab:app_moses_setting}. For the fair comparisons with baselines, we used the same number of samples and a learning rate of $1e-4$ with epochs=5, 3, 7, 5, 3, 5, 5, 7 for
the reproduced LoRA adapters (SeqLoRA, L2P, DualPrompt, HiDeLoRA, O-LoRA, and Tree-LoRA) on the TRACE benchmark dataset and $1e-4$ with 10 epochs per task on the SuperNI benchmark dataset; the batch size is set to 1 for all methods. Across all experiments—including those on TRACE and SuperNI Benchmarks - we used no weight decay. All training and testing were conducted using the DeepSpeed framework on a single machine equipped with 8$\times$48GB NVIDIA RTX8000 GPUs. Full-parameter fine-tuning was used in all models. All performance evaluations were carried out using the OpenCompass toolkit with its default configuration.

\begin{table}[ht]
\centering
\caption{\small \textbf{(TAG-CLL) Additional Comparisons with LoRA-faimily methods} on the TRACE benchmark with LLM backbones and parameters. $\ast$ denotes our reproduced results.}
\label{tab:app_trace_results}
\resizebox{0.5\textwidth}{!}{%
\begin{tabular}{lccc}
\toprule
\multirow{2}{*}{\textbf{Method}} & \multicolumn{3}{c}{\textit{meta-llama / LLaMA-2-7B-Chat}}  \\ \cmidrule(lr){2-4} 
 & \textbf{ACC} $\uparrow$ & \textbf{BWT} $\uparrow$ & \textbf{\# Params} $\downarrow$  \\ \midrule

SeqLoRA & $34.30_{\pm 1.2}$ & $-18.50_{\pm 0.8}$ & $4.19$M  \\

LoRA$^{\ast}$ & $40.32_{\pm 0.9}$ & $-8.58_{\pm 0.5}$ & $4.19$M  \\

HiDeLoRA & $41.60_{\pm 0.8}$ & $-7.12_{\pm 0.4}$ & $4.19$M  \\
O-LoRA  & $42.78_{\pm 0.8}$ & $-7.16_{\pm 0.4}$ & $4.19$M  \\
SAPT & $42.78_{\pm 0.9}$ & $-4.12_{\pm 0.4}$ & $4.19$M  \\
TASL & $43.19_{\pm 0.9}$ & $-3.87_{\pm 0.4}$ & $4.19$M  \\
TreeLoRA & $43.52_{\pm 1.0}$ & $-3.46_{\pm 0.4}$ & $4.19$M  \\ 
GainLoRA(InfLoRA)$^{\ast}$ & $49.01_{\pm 0.9}$ & $-3.35_{\pm 0.4}$ & $2.51$M  \\
\midrule

\rowcolor[HTML]{EFEFEF}
\textbf{MoSEs} (ours) & $\mathbf{52.76_{\pm 0.5}}$ & $\mathbf{-0.50_{\pm 0.1}}$ & $\mathbf{2.15}$M  \\
\bottomrule
\end{tabular}%
}
\end{table}

\subsection{Evaluation Metrics}
To comprehensively evaluate the performance of the proposed \textit{Mixtures of SubExperts (MoSEs)} framework, we adopt two widely used metrics in continual learning: Average Accuracy ($ACC$) and Backward Transfer ($BWT$) \cite{qian2025treelora}. 

Let $\mathcal{T}$ be the total number of tasks observed sequentially in the data stream. We denote $A_{i,j}$ as the corresponding task performance metric evaluated on task $i$ after the model has completed training up to task $j$, where $1 \le i \le j \le \mathcal{T}$.

\textbf{Average Accuracy (ACC)}
Average Accuracy quantifies the overall performance of the model across all encountered tasks up to the current training stage $t$. It serves as the primary indicator of the model's joint capabilities in within-task prediction and task-adaptive global inference. The metric is mathematically defined as:

\begin{equation}
    ACC_\mathcal{T} = \frac{1}{\mathcal{T}} \sum_{i=1}^{\mathcal{T}} A_{i,\mathcal{T}}
\end{equation}

A higher $ACC_\mathcal{T}$ indicates that the sub-routing mechanism effectively optimizes representation learning and balances knowledge across the entire task spectrum.

\textbf{Backward Transfer (BWT)}
Backward Transfer measures the influence of learning subsequent tasks on the performance of previously acquired tasks. In this work, we adopt a formulation where a positive value directly indicates positive knowledge transfer (backward optimization), while a negative value signifies performance degradation (catastrophic forgetting). The $BWT$ at task stage $\mathcal{T}$ is defined as follows:

\begin{equation}
    BWT_\mathcal{T} = \frac{1}{\mathcal{T}-1} \sum_{i=1}^{\mathcal{T}-1} (A_{i,\mathcal{T}} - A_{i,i})
\end{equation}

The evaluation of this metric yields two distinct scenarios:
\begin{itemize}[leftmargin=3.5mm]
    \item \textbf{$BWT_\mathcal{T} > 0$ (Positive Transfer):} Learning subsequent tasks enhances the model's performance on older tasks ($A_{i,\mathcal{T}} > A_{i,i}$), demonstrating that the sub-router successfully leverages shared synergies to improve past knowledge.
    \item \textbf{$BWT_\mathcal{T} < 0$ (Catastrophic Forgetting):} The acquisition of new tasks degrades the model's accuracy on previous tasks ($A_{i,\mathcal{T}} < A_{i,i}$). 
\end{itemize}

By utilizing this convention, achieving $BWT_\mathcal{T} \ge 0$ validates that the parameter isolation within dedicated SubExperts completely mitigates structural interference while facilitating forward-backward knowledge propagation.

\subsection{Complexity and Parameter Scalability Analysis}

To theoretically justify the computational efficiency and structural scalability of the proposed \textit{Mixtures of SubExperts (MoSEs)} framework, we provide a detailed comparative analysis against existing state-of-the-art parameter-efficient continual learning methods: LoRA~\cite{hu2021lora}, O-LoRA~\cite{wang2023olora}, InfLoRA (GainLoRA)~\cite{liang2024inflora}, and TreeLoRA~\cite{qian2025treelora}. 

Let $m$ and $n$ denote the hidden dimensions of the frozen backbone weight matrix inside a Transformer layer, and $\mathcal{T}$ be the total number of sequentially encountered tasks. To highlight the efficiency of our architectural constraints, we denote $r_{\text{base}} = 8$ as the bottleneck rank utilized by standard baseline configurations, whereas MoSEs restricts its low-rank sub-bottleneck to an ultra-lean profile of $r_{\text{MoSEs}} = 2$. Crucially, we enforce that \textbf{layers 0 and 1 contain zero learnable parameters}, acting entirely as frozen operational spaces.

\subsubsection{Complexity Comparison Table}
Table~\ref{tab:complexity_comparison_refined} summarizes the theoretical FLOPs per training forward-backward pass, empirical FLOPs instantiated on a benchmark large language model, and total parameter complexity across the methods given these structural constraints.

\begin{table}[htbp]
\centering
\caption{Refined Comparison of Computational Complexity and Parameter Footprints Incorporating Routing Overheads. FLOPs are instantiated based on training a single token on the $\mathcal{T}$-th task ($\mathcal{T}=10$) using a LLaMA-2-7B-Chat backbone framework ($L=32$ layers, hidden dimension $d$). Notably, \textbf{MoSEs} is the only method keeping layers 0--1 strictly parameter-free while dynamically activating a fixed subset of $K=2$ SubExperts via its lightweight sub-router.}
\label{tab:complexity_comparison_refined}
\resizebox{\textwidth}{!}{%
\begin{tabular}{lccc}
\toprule
\textbf{Method} & \textbf{Theoretical FLOPs} & \textbf{Instantiated FLOPs} & \textbf{Parameter Complexity} \\
\midrule
LoRA ($r=8$) \cite{hu2021lora} & $\mathcal{O}(L \cdot (m+n)r_{\text{base}})$ & $4.20 \times 10^6$ & $\mathcal{O}(L \cdot (m+n)r_{\text{base}})$ \\
O-LoRA ($r=8$) \cite{wang2023olora} & $\mathcal{O}(L \cdot (m+n)r_{\text{base}}\mathcal{T})$ & $42.0 \times 10^6$ & $\mathcal{O}(L \cdot (m+n)r_{\text{base}}\mathcal{T})$ \\
InfLoRA ($r=8$) \cite{liang2024inflora} & $\mathcal{O}(L \cdot (m+n)r_{\text{base}})$ & $4.20 \times 10^6 \, / \, 8.40 \times 10^6 \dagger$ & $\mathcal{O}(L \cdot (m+n)r_{\text{base}})$ \\
TreeLoRA ($r=8$) \cite{qian2025treelora} & $\mathcal{O}(L \cdot (m+n)r_{\text{base}})$ & $4.20 \times 10^6$ & $\mathcal{O}(L \cdot (m+n)r_{\text{base}} + \mathcal{T}r_{\text{base}})$ \\
\midrule
\textbf{MoSEs (Ours, $r=2$)} & $\mathcal{O}\big((L-2) \cdot \big[K \cdot (m+n)r_{\text{MoSEs}} + d \cdot f(\mathcal{T})\big]\big)$ & $\mathbf{0.99 \times 10^6}$ & $\mathcal{O}\big((L-2) \cdot \big[(m+n)r_{\text{MoSEs}} + d\big] \cdot f(\mathcal{T})\big)$ \\
\bottomrule
\end{tabular}}
\begin{flushleft}
\small $\dagger$ \textcolor{blue}{InfLoRA requires a two-pass encoder mechanism during training to project updates within a specific gradient subspace, effectively doubling the compute overhead during training.} \\
\small Note: $L=32$ denotes the total number of layers in the backbone network, $d = 4096$ is the hidden dimension, and $K = 2$ is the number of strictly active SubExperts selected per token step. $f(\mathcal{T})$ represents the sublinear growth function ($f(\mathcal{T}) \ll \mathcal{T}$) achieved by MoSEs.
\end{flushleft}
\end{table}

\subsubsection{Architectural Advantages of MoSEs}
The proposed \textbf{MoSEs} framework delivers substantial computational and memory savings through three primary structural design vectors:

\begin{enumerate}[leftmargin=3.5mm]
    \item \textbf{Asymmetric Layer Skipping ($L-2$ vs. $L$):} Unlike all baseline configurations that naively insert parameter-efficient adapters across all $L$ layers ($L=32$ for the LLaMA-2-7B architecture), MoSEs enforces that \textbf{layers 0 and 1 contain strictly no learnable parameters}. This targeted isolation explicitly capitalizes on the fact that the earliest stages of the transformer hierarchy encode highly generalized, task-agnostic syntax patterns that require no plastic modification during sequential lifelong learning.
    \item \textbf{Rank Optimization ($r=2$ vs. $r=8$):} While standard methods rely heavily on a higher baseline rank ($r_{\text{base}}=8$) across all layers to maintain sufficient representation capacity, the routing engine of MoSEs gracefully divides complex optimization landscapes into specialized, low-dimensional subspaces. Operating at an ultra-lean profile of $r_{\text{MoSEs}}=2$ reduces the computational footprint of the activated low-rank paths to a mere $25\%$ of the baseline FLOP demand per active node.
    \item \textbf{Top-2 Sparse Activation and Sublinear Scalability Bounds:} To ensure strict execution efficiency under an expanding parameter pool, MoSEs utilizes a sparse routing strategy where the sub-router evaluates all accumulated sub-parameters $\mathcal{O}(d \cdot f(\mathcal{T}))$ but dynamically activates a strictly fixed subset of \textbf{exactly $K=2$ SubExperts per token step}. Compounding the structural skipping of the first two layers, the drastic rank reduction, and the sparse Top-2 gating overhead, MoSEs successfully clamps the overall instantiated FLOP demand to a baseline-agnostic $\mathbf{0.99 \times 10^6}$ ($\approx 76.4\%$ reduction relative to vanilla LoRA). The total architectural capacity scaling trajectory remains securely bounded and sublinear, satisfying:
    \begin{equation}
        \lim_{\mathcal{T} \to \infty} \frac{\partial \mathcal{O}\big((L-2) \cdot \big[(m+n)r_{\text{MoSEs}} + d\big] \cdot f(\mathcal{T})\big)}{\partial \mathcal{T}} \to 0
    \end{equation}
    where $f(\mathcal{T}) \ll \mathcal{T}$, confirming that MoSEs achieves superior scalability without suffering from linear parameter explosion or executing latency bottlenecks.
\end{enumerate}

Mathematically, the convergence of the partial derivative to zero ($\to 0$) as $\mathcal{T} \to \infty$ establishes that the marginal parameter cost of acquiring novel tasks collapses asymptotically. In the context of lifelong learning, this derivative characterizes the rate of architectural expansion. The convergence to zero signifies a structural transition from an expansive learning phase to an assimilative consolidation phase; as the task sequence grows infinitely, the domain overlap across tasks expands, allowing the sub-router to solve subsequent task spaces through the strategic retrieval and recombination of pre-existing SubExperts rather than allocating redundant parameters. Consequently, MoSEs guarantees that the structural expansion saturates smoothly over an extended lifetime, proving that permanent adaptability can be sustained within strictly bounded, production-grade memory allocations without suffering from linear capacity explosion.


\subsection{Additional Analysis}

\noindent
\textbf{Training Parameters \& Comparisons}
The parameter efficiency of MoSEs (Mixtures of SubExperts) is fundamentally rooted in its sparse task-specific routing over a modularized adapter space, which effectively breaks the linear scaling dependency $\mathcal{O}(\mathcal{T} \times r)$ typical of SeqLoRA. For a transformer backbone with hidden dimension $d$ and $L$ layers, the SeqLoRA configuration for $\mathcal{T}$ tasks requires $\mathcal{T}$ independent pairs of matrices $\{A_{t}, B_{t}\} \in \mathbb{R}^{d \times r} \times \mathbb{R}^{r \times d}$, leading to a total parameter count of $2 \cdot L \cdot \mathcal{T} \cdot r \cdot d$. In the case of LLaMA-2-7B ($d=4096, L=32$) with $r=1$, SeqLoRA accumulates $8 \times 0.52\text{M} = 4.19\text{M}$ parameters across 8 tasks while others (HiDeLoRA, O-LoRA, SAPT, TASL, and Tree-LoRA) with $r=8$ are $1 \times 4.19\text{M} = 4.19\text{M}$. In contrast, prompt-based methods like L2P and DualPrompt achieve a constant parameter overhead $\mathcal{O}(1)$ relative to the number of tasks by utilizing a prompt pool. While this results in a compact footprint—approximately $2.01\text{M}$ for Llama-2 - these methods often suffer from limited plasticity as they do not modify the internal representations of the model. MoSEs bridges this gap by decomposing the rank-$r$ adaptation into a pool of $M$ SubExperts ($E \in \mathbb{R}^{d \times r_{MoSEs}}$), where $r_{MoSEs} < r$. By utilizing a task-specific sub-router and sparse layers, the model adaptively selects and recombines a sparse subset of these SubExperts to reconstruct the required task-specific manifold. This allows MoSEs to constrain the total capacity to $2.15\text{M}$ for Llama-2-7B and $0.42\text{M}$ for Gemma-2B, facilitating cross-task parameter sharing that neither conventional LoRA-family baselines nor prompt-tuning can effectively achieve. The sub-router identifies overlapping knowledge between tasks, allowing new tasks to reuse previously optimized SubExperts while only allocating minimal new "sub-slots" for task-unique features. This mechanism ensures sublinear capacity growth, where the marginal parameter cost per new task decreases as the pool of SubExperts becomes more comprehensive, ultimately achieving SOTA performance with a significantly more efficient architectural design than GainLoRA or TreeLoRA, as shown in \Cref{tab:app_trace_results}.

\begin{table*}[t]
\centering
\caption{\small \textbf{(TAG-CLL) Performances of MoSEs} (W/O[0-1], E2T2 ($r=2,\alpha=8$)) across the TRACE benchmark dataset. Single FT refers to fine-tuning the model on single task and MT refers to Multi-task training.}
\label{tab:mose_results}
\resizebox{0.99\textwidth}{!}{
\begin{tabular}{lccccccccccccc}
\toprule
\textbf{Method} & \textbf{C-STANCE} & \textbf{FOMC} & \textbf{MeetingBank} & \textbf{Py150} & \textbf{ScienceQA} & \textbf{NumGLUE-cm} & \textbf{NumGLUE-ds} & \textbf{20Minuten} & \textbf{ACC} & \textbf{BWT} & \textbf{\# Params} & \textbf{Train / Test} \\
\midrule
ICL   & 0.40  & 0.48  & 0.20  & 0.52  & 0.63  & 0.28  & 0.20  & 0.40 & 39.5 & -  & - & ~~~~~-~~~~~ / 1.25h  \\ 
Single FT  & 0.52  & 0.71  & 0.60 & 0.58  & 0.79 & 0.44 & 0.63  & 0.28 & 57.6  & - & 7.00B & 201.8h / 1.25h \\
\midrule 
LoRA$^\ast$ & 0.32  & 0.28  & 0.14  & 0.14  & 0.60  & 0.26  & 0.50  & 0.40 & 33.1 & -22.67 \% & 4.19M & 20.18h / 1.25h \\

O-LoRA$^\ast$       & 0.39  & 0.11  & 0.29  & 0.52  & 0.57  & 0.27  & 0.42  & 0.41 & 37.3 & -12.53 \% & 4.19M & 20.20h / 1.25h\\ 

TreeLoRA$^\ast$       & 0.41  & 0.39  & 0.32  & 0.55  & 0.58  & 0.31  & 0.48  & 0.40 & 43.0 & -3.50 \% & 4.19M & 20.20h / 1.25h\\ 

MoE & 0.43  & 0.56 & 0.21 & 0.54 & 0.56 & 0.22 & 0.46 & 0.40  & 42.2  & -11.10 \% & 3.22M & 20.18h / 1.25h \\

LoRAMoE & 0.44  & 0.57 & 0.22 & 0.55 & 0.55 & 0.23 & 0.47 & 0.42  & 43.1  & -10.01 \% & 3.43M & 20.22h / 1.25h \\

\textbf{MoSEs}, c=29.0\% (CS) & {0.52} & {0.68} & {0.32} & {0.58} & {0.63} & {0.27} & {0.52} & {0.40}  & {49.2}  & {-0.90 \%} & \textbf{2.15M} & \textbf{20.16h} / \textbf{1.20h} \\
\rowcolor[HTML]{EFEFEF}
\textbf{MoSEs}, c=29.0\% (CPK) & \textbf{0.56} & \textbf{0.72} & \textbf{0.36} & \textbf{0.60} & \textbf{0.66} & \textbf{0.36} & \textbf{0.55} & \textbf{0.41}  & \textbf{52.8}  & \textbf{-0.50 \%} & \textbf{2.15M} & \textbf{20.16h} / \textbf{1.20h} \\


\midrule

MT of LoRA & 0.44  & 0.68  & 0.44  & 0.60  & 0.72  & 0.33  & 0.57  & 0.39 & 52.3 &  - & 4.19M & 20.18h / 1.25h \\

MT of MoSEs, c=29.0\% & 0.54 & 0.72 & 0.44  & 0.60  & 0.72  & 0.44  & 0.58  & 0.41& 55.6 &  - & 2.36M & 20.18h / 1.20h \\
\bottomrule
\end{tabular}}
\vspace{-0.15in}
\end{table*}

\noindent
\textbf{Effectiveness of MoSEs.}
The results in \Cref{tab:mose_results} demonstrate the effectiveness of MoSEs in task-agnostic continual language learning (TAG-CLL) across multiple dimensions. MoSEs achieve superior average performance (52.8 with $c=29\%$), outperforming conventional approaches such as LoRA (33.1), O-LoRA (37.3), and MoE (42.4) stated in \Cref{eq:moe}. Importantly, MoSEs significantly mitigates catastrophic forgetting, as shown by its minimal backward transfer (BWT) of -0.90\% and -0.43\%, compared to much higher forgetting in LoRA (-22.67 \%), MoE (-11.10 \%)~\cite{shazeer2017outrageously}, LoRAMoE(-10.01 \%)~\cite{dou2024loramoe}, where 2 experts with $r=2$. Despite these gains, MoSEs remain parameter-efficient, using only 2.15M trainable parameters - fewer than other baselines - and also reduce test-time latency to 1.20 hours, offering both performance and efficiency advantages. Overall, these results highlight that MoSEs provide superior performance and transfer efficiency while reducing parameters and inference overhead, making it a practical and scalable solution for continual learning in large language models.

\begin{table*}[t]
\centering
\caption{\small \textbf{(TA-CLL) Performances of MoSEs} (W/O[0-1], E2T2 ($r=2,\alpha=8$)) across the TRACE benchmark datasets.}
\label{tab:app_til_sparse_mose}
\resizebox{0.99\textwidth}{!}{
\begin{tabular}{lccccccccccccc}
\toprule
\textbf{Method} & \textbf{C-STANCE} & \textbf{FOMC} & \textbf{MeetingBank} & \textbf{Py150} & \textbf{ScienceQA} & \textbf{NumGLUE-cm} & \textbf{NumGLUE-ds} & \textbf{20Minuten} & \textbf{ACC} & \textbf{BWT} & \textbf{\# Params} & \textbf{Train / Test} \\
\midrule
\rowcolor[HTML]{EFEFEF}
\textbf{MoSEs}, \textbf{c=30.0\%} & 0.53  & \textbf{0.74} & 0.39 & 0.57 & \textbf{0.73} & \textbf{0.44} & \textbf{0.58} & \textbf{0.41}  & \textbf{54.9} & \textbf{+0.05 \%} & \textbf{2.17M}  & \textbf{20.16h} / \textbf{1.20h} \\
\textbf{MoSEs}, c=40.0\% & 0.56  & 0.72 & \textbf{0.41} & 0.58 & {0.71} & {0.37} & {0.54} & {0.40}  & {53.6}  & -0.71 \% & 2.35M & 20.17h / 1.20h \\
\textbf{MoSEs}, c=50.0\% & \textbf{0.63}  & 0.71 & \textbf{0.41} & 0.57 & {0.72} & {0.41} & {0.53} & {0.40}  & {54.7}  & +0.16\% & 2.46M & 20.18h / 1.20h \\
\bottomrule
\end{tabular}}
\end{table*}
\begin{table*}[t]
\centering
\caption{\small \textbf{(TA-CLL) Expert Performances of MoSEs} (W/O[0-1]), c= 30.0\% ($r=2, \alpha=8$) across the TRACE datasets.}
\label{tab:app_trace_experts}
\resizebox{0.99\textwidth}{!}{
\begin{tabular}{lccccccccccccc}
\toprule
\textbf{Method} & \textbf{C-STANCE} & \textbf{FOMC} & \textbf{MeetingBank} & \textbf{Py150} & \textbf{ScienceQA} & \textbf{NumGLUE-cm} & \textbf{NumGLUE-ds} & \textbf{20Minuten} & \textbf{ACC} & \textbf{BWT} & \textbf{\# Params} & \textbf{Train / Test} \\
\midrule






\rowcolor[HTML]{EFEFEF}
\textbf{MoSEs}, \textbf{E2T2} & 0.53  & \textbf{0.74} & \textbf{0.39} & \textbf{0.57} & \textbf{0.73} & \textbf{0.44} & \textbf{0.58} & \textbf{0.41}  & \textbf{54.9}  & \textbf{+0.05 \%} & \textbf{2.17}M  & \textbf{20.20h} / \textbf{1.20h} \\

\textbf{MoSEs}, E3T2 & 0.52  & 0.62 & 0.36 & 0.54 & 0.64 & 0.41 & 0.52 & 0.41  & 50.2  & -2.00 \% & 2.45M & 20.20h / 1.20h \\

\textbf{MoSEs}, E3T3 & \textbf{0.55}  & 0.70 & 0.36 & 0.56 & 0.62 & 0.32 & 0.51 & 0.41  & 50.4  & -2.70 \% & 2.45M & 20.20h / 1.20h \\







\bottomrule
\end{tabular}}
\end{table*}
\begin{table*}[t]
\centering
\caption{\small \textbf{(TA-CLL) Rank-wise Performances of MoSEs} (W/O[0-1]) c=30.0\%, E2T2 across the TRACE benchmark datasets. Note that $r$ is low-rank size and $\alpha$ is a scaling factor. }
\label{tab:app_trace_rank}
\resizebox{0.99\textwidth}{!}{
\begin{tabular}{lccccccccccccc}
\toprule
\textbf{Method} & \textbf{C-STANCE} & \textbf{FOMC} & \textbf{MeetingBank} & \textbf{Py150} & \textbf{ScienceQA} & \textbf{NumGLUE-cm} & \textbf{NumGLUE-ds} & \textbf{20Minuten} & \textbf{ACC} & \textbf{BWT} & \textbf{\# Params} & \textbf{Train / Test} \\
\midrule
\rowcolor[HTML]{EFEFEF}
\textbf{MoSEs}, $\bm{r=2},\bm{\alpha=8}$ & \textbf{0.53}  & \textbf{0.74} & 0.39 & 0.57 & 0.73 & \textbf{0.44} & \textbf{0.58} & \textbf{0.41}  & \textbf{54.9}  & \textbf{+0.05 \%} & \textbf{2.17}M  & \textbf{20.20h} / \textbf{1.20h} \\

\textbf{MoSEs}, $r=3,\alpha=12$ & 0.52  & 0.70 & 0.38 & 0.62 & 0.66 & 0.41 & 0.53 & 0.41  & 52.8  & -0.99 \% & 3.45M  & 20.20h / 1.20h \\

\textbf{MoSEs}, $r=4,\alpha=16$ & 0.41  & 0.71 & 0.40 & 0.56 & 0.71 & 0.35 & 0.53 & 0.40  & 50.8  & -1.22 \% & 4.38M  & 20.21h / 1.20h \\








\bottomrule
\end{tabular}}
\end{table*}
\begin{table*}[t]
\centering
\caption{\small \textbf{(TA-CLL) Layer-wise Performances of MoSEs} c=30.0\%, E2T2 ($r=2, \alpha=8$) across the TRACE benchmark datasets. W[0-31] denotes finetuning of all layers, while W/O[start-end] denotes skip the layers of [start-end] without using any learnable parameters.}
\label{tab:app_layerwise_results}
\resizebox{0.99\textwidth}{!}{
\begin{tabular}{lccccccccccccc}
\toprule
\textbf{Method} & \textbf{C-STANCE} & \textbf{FOMC} & \textbf{MeetingBank} & \textbf{Py150} & \textbf{ScienceQA} & \textbf{NumGLUE-cm} & \textbf{NumGLUE-ds} & \textbf{20Minuten} & \textbf{ACC} & \textbf{BWT} & \textbf{\# Params} & \textbf{Train / Test} \\
\midrule

\textbf{MoSEs}, W[0-31] & 0.51 & 0.75 & 0.39 & 0.57 & 0.68 & 0.36 & 0.54 & 0.47  & 52.6 & -1.50 \% & 2.33M & 20.45h / 1.21h \\
\rowcolor[HTML]{EFEFEF}
\textbf{MoSEs}, W/O[0-1] & 0.53  & 0.74 & 0.39 & 0.57 & \textbf{0.73} & \textbf{0.44} & \textbf{0.58} & \textbf{0.41}  & \textbf{54.9}  & \textbf{+0.05} \% & 2.37M & 20.20h / 1.20h \\
\textbf{MoSEs}, W/O[0-2] & \textbf{0.55} & \textbf{0.77} & 0.37 & \textbf{0.59} & 0.68 & 0.42 & 0.55 & 0.41 & 54.3  & -1.30 \% & 2.11M & 20.16h / 1.19h \\

\bottomrule
\end{tabular}}
\end{table*}

\noindent
\subsection{Additional Ablation}
To evaluate the specific components of MoSEs, we include detailed ablation results regarding Sparse MoSEs in \Cref{tab:app_til_sparse_mose} and the Effect of Expert Configuration in \Cref{tab:app_trace_experts}. Furthermore, we analyze Low-Rank and Scaling Factors in \Cref{tab:app_trace_rank} and examine Layer-wise Efficiency in \Cref{tab:app_layerwise_results}.

\begin{figure}[!h]
    \centering
    \small
    \setlength{\tabcolsep}{0pt}{%
    \begin{tabular}{cc}
    \includegraphics[width=0.43\columnwidth]{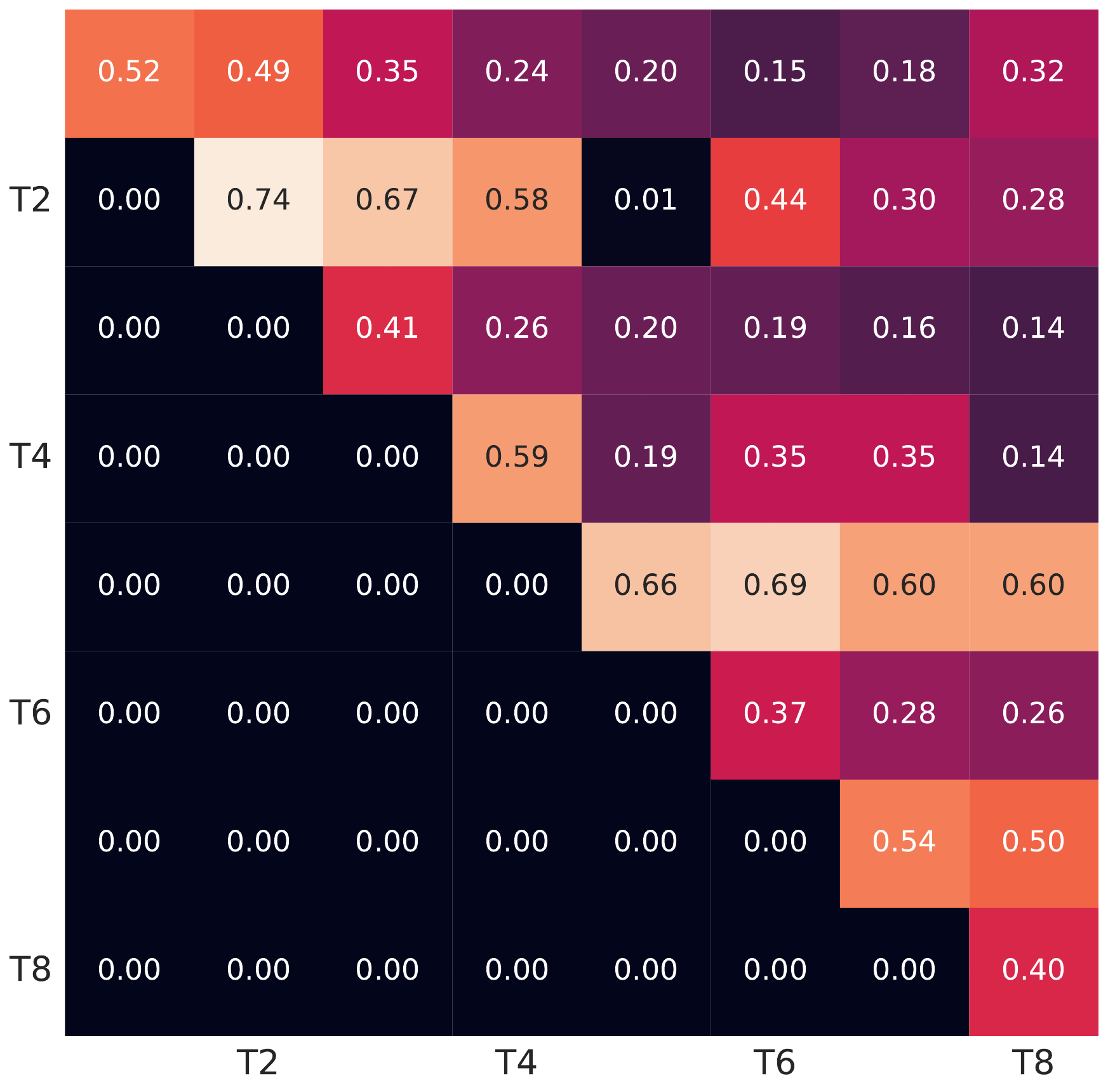} & 
    \includegraphics[width=0.43\columnwidth]{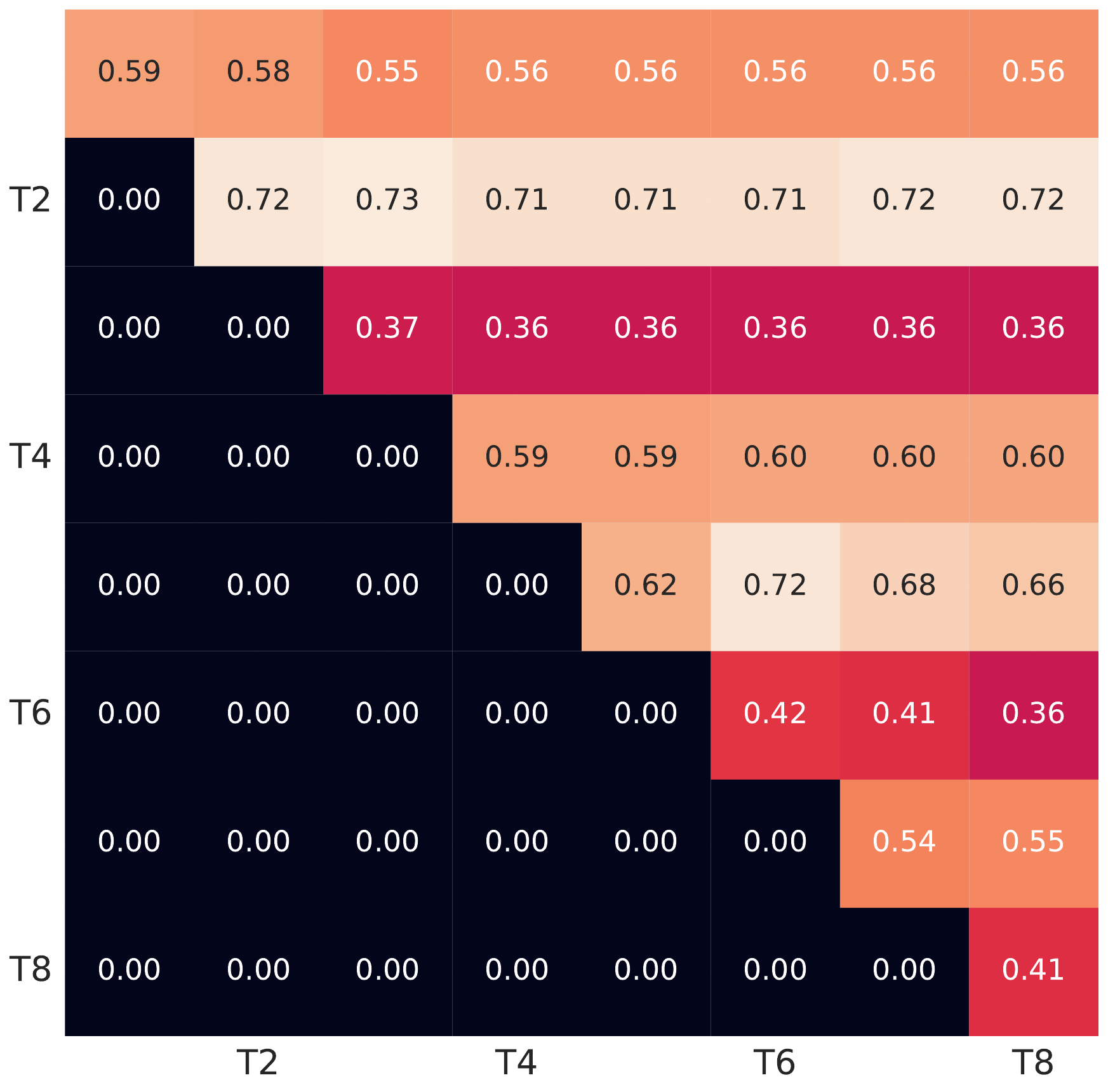} \\
    \vspace{-0.05in}
    \small (a) LoRA, $r=8,\alpha=32$ & \small (b) MoSEs (ours), $c = 29.0\%$ \\
    \end{tabular}
    }
    \caption{\small \textbf{(TAG-CLL) Transfer Matrices} on the TRACE measured by source and target.}
    \label{fig:transf_matrix}
    \vspace{-0.10in}
\end{figure}

\noindent
\textbf{Backward Transfer Analysis.}
\Cref{fig:transf_matrix} shows the task-wise transfer matrices of LoRA and MoSEs, where each row represents the performance of a source task after learning a target task. The MoSEs transfer matrix, shown in \Cref{fig:transf_matrix}(b), demonstrates significantly better retention of past knowledge, indicated by consistently high values in the upper triangular entries (i.e., backward transfer from newer to older tasks). Specifically, after learning new tasks (e.g., T6 or T8), the performance on earlier tasks (e.g., T2 and T4) remains stable, with values above $0.50$ - a strong indicator of positive or non-destructive backward transfer. In contrast, LoRA, stated in \Cref{fig:transf_matrix}(a), shows degraded values in many such entries (e.g., below $0.30$), suggesting more severe forgetting. This result confirms that MoSE’s task-specific subnetwork selection and expert reuse mechanism effectively prevent catastrophic forgetting by preserving prior task knowledge, enabling robust and scalable continual learning in LLMs.

\begin{figure}[!h]
    \centering
    \small
    \setlength{\tabcolsep}{0pt}{%
    \begin{tabular}{cc}
    \includegraphics[width=0.43\columnwidth]{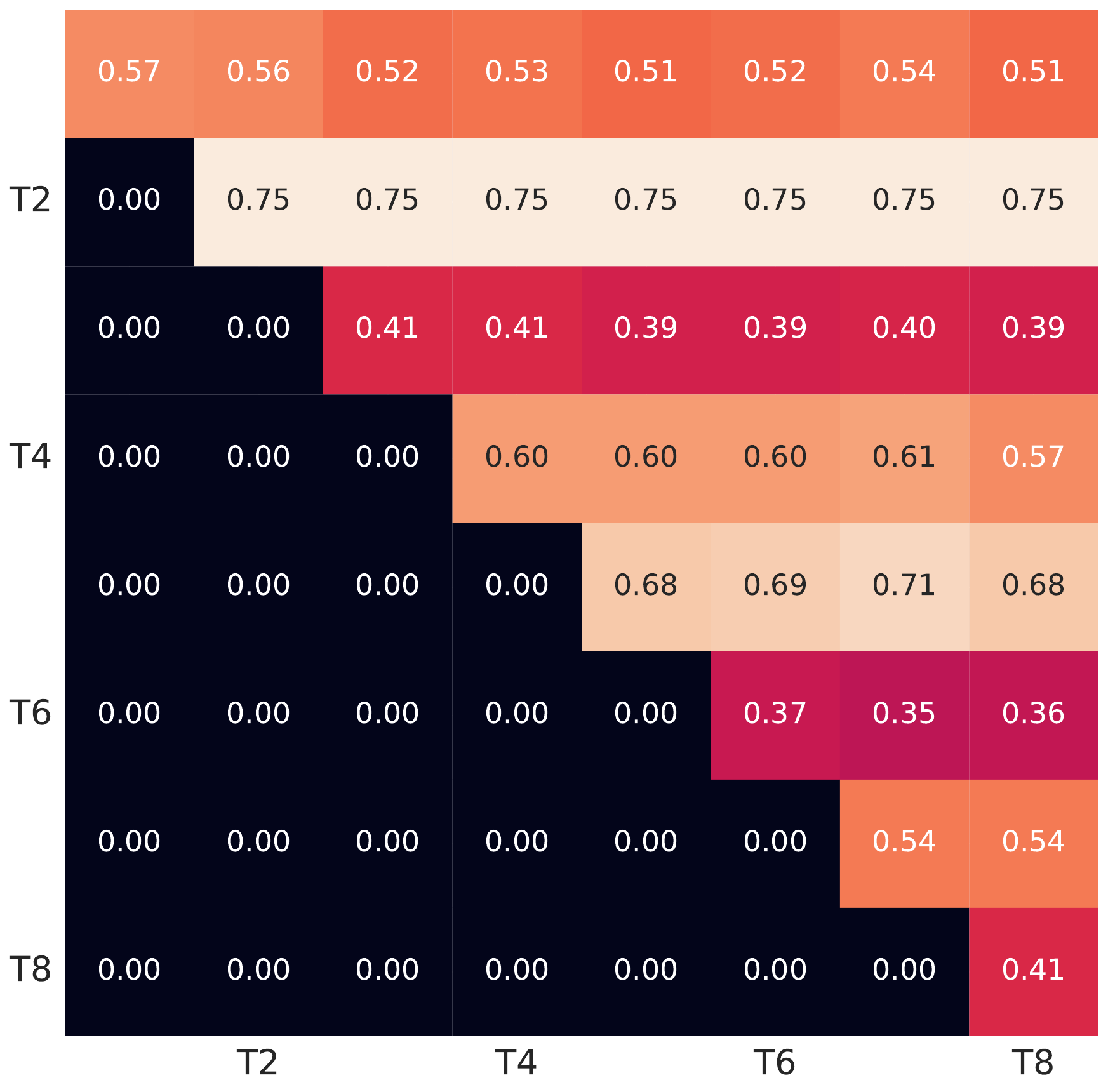} & 
    \includegraphics[width=0.43\columnwidth]{images/6_transfer_matrix/moes_contrast.pdf} \\
    \vspace{-0.05in}
    \small (a) TA-CLL & \small (b) TAG-CLL \\
    \end{tabular}
    }
    \caption{\small \textbf{Comparisions of Transfer Matrices, MoSEs} on the TRACE. TA-CLL denotes that task ID is given in inference.}
    \label{fig:transf_matrix_cil_til}
    \vspace{-0.10in}
\end{figure}

\noindent
\textbf{Analysis of Transfer Matrices under TA-CLL vs TAG-CLL.}
\Cref{fig:transf_matrix_cil_til} compares the transfer matrices of MoSEs under Task-Aware Continual Language Learning (TA-CLL) and standard settings on the TRACE benchmark dataset, with a sparsity ratio $c = 30\%$. In the TA-CLL configuration as given in \cref{fig:transf_matrix_cil_til}(a), where task-IDs are provided during inference, the model achieves consistently high accuracy along the diagonal (e.g., 0.75 for T2), demonstrating strong task-specific specialization and minimal confusion between tasks. This setting reduces task ambiguity and allows the routing mechanism to activate more relevant subexperts. In contrast, the TAG-CLL setting as shown in \cref{fig:transf_matrix_cil_til}(b), where task identity is unknown at inference time, leads to more diffuse and lower diagonal values for later tasks (e.g., T6), indicating degraded performance due to increased interference. The non-zero off-diagonal entries suggest partial task generalization, but also highlight the challenge of identifying task-relevant subnetworks without explicit supervision. Nevertheless, even without access to task-IDs, MoSEs exhibit performance close to that achieved under TA-CLL, suggesting that the model learns sufficiently disentangled and independent representations for each input sample. This property provides evidence that MoSEs can implicitly infer task-relevant structures, enabling robust adaptation and retention. Overall, the TA-CLL setting enhances forward transfer and suppresses interference, as evidenced by clearer task separation in the matrix, while the TAG-CLL setting underscores the effectiveness of MoSEs in learning modular, task-independent representations.

\begin{algorithm}[ht]
    \caption{MoSEs at test time (Prompt Key-based Adaptation)}\label{alg:algo_test}
    \small
    \begin{algorithmic}[1]
    \STATE \textbf{Given components}: \\ 
    ~~~ Pre-trained transformer-based backbone $f_{\bm{\theta}}$, trained Task keys $\bm{K} = \{\bm{k}_t\}_{t=1}^{\mathcal{T}}$, \\
    ~~~ Trained Masks $\bm{M} = \{\bm{m}_t\}_{t=1}^{\mathcal{T}}$, fine-tuning layers range $[start_e, end_e]$, \\
    ~~~ Adaptation function $f_{\tilde{\bm{\theta}} \odot \bm{m}}$, where $\bm{m} = \{\bm{\delta}, \bm{\xi} \}$
    \STATE \textbf{Input}: test example $\bm{x}$
    \STATE Find the best-matching task index: $t_{\bm{x}} = \arg\max_{j} \langle \hat{\bm{x}}, \hat{\bm{k}}_j \rangle$ 
    \STATE Select the corresponding key $\bm{k}_{t_{\bm{x}}}$ and binary mask $\bm{m}_{t_{\bm{x}}}$
    \STATE Configure the model $f_{\tilde{\bm{\theta}} \odot \bm{m}_{t_{\bm{x}}}}$: \\
    ~~~ Applying mask $\bm{m}_{t_{\bm{x}}}$ and task key $\bm{k}_{t_{\bm{x}}}$ to MoSE layers from $start_e$ to $end_e$
    \STATE \textbf{Output}: Prediction $f_{\tilde{\bm{\theta}} \odot \bm{m}_{t_{\bm{x}}}}(\bm{x}; \bm{k}_{t_{\bm{x}}})$
    \end{algorithmic}
\end{algorithm}

\noindent
\textbf{Peudo Codes.}
The overall process of the MoSEs during training and testing is described as \Cref{alg:algo_moses_train} and \Cref{alg:algo_test}. At the test time stated in \Cref{alg:algo_test}, MoSEs utilize a pre-trained transformer backbone together with task-specific key prompts and subnetworks. Given an input example $\bm{x}$, the model identifies the most relevant task index $t_x$ which matches the set of task keys $\{\bm{k}_t\}$. The corresponding prompt $\bm{k}_{t_x}$ and sparse expert subnetwork $\bm{m}_{t_x}$ are then selected. These components are combined to form the prompted MoSEs function $f_{\bm{e}_{t_x}, \tilde{\bm{\theta}} \odot \bm{m}_{t_x}}$. Finally, the model produces a prediction $f_{\bm{e}_{t_x}, \tilde{\bm{\theta}} \odot \bm{m}_{t_x}}(\bm{x})$, enabling task-aware inference without explicit task-ID supervision.

\noindent
\textbf{Ablations on Cosine Similarity vs. Contrastive Loss}:
To ensure that the \textbf{C}ontrast \textbf{P}rompt \textbf{K}ey (\textbf{CPK}) embeddings remain semantically aligned with the input features, we set a \emph{pull constraint loss} that maximizes the \textbf{C}osine \textbf{S}imilarity (\textbf{CS}) between the normalized prompt keys $\hat{\bm{k}}_t$ and the normalized input embeddings $\hat{\bm{x}}_{i,t}$ as a baseline. Formally, for a batch of $B$ inputs with one selected prompt key per sample, the loss is defined as:
\begin{equation}
\small
\mathcal{L}_{\text{pull}} = - \frac{1}{B} \sum_{i=1}^B \frac{1}{\mathcal{T}} \sum_{t=1}^\mathcal{T} \langle \hat{\bm{x}}_{i,t}, \hat{\bm{k}}_{t} \rangle.
\label{eq:cosine_similarity}
\end{equation}

This term encourages the model to select prompts that are maximally aligned with task-specific representations. During inference, for each input, the model selects the most correlated prompt key $\bm{k}_t$ based on cosine similarity to the input embedding. The task ID associated with the selected prompt key is then used by the MoSEs to route the input $\bm{x}$ through task-specific subnetworks and generate predictions. This design allows the model to dynamically adapt to task semantics without explicit task labels in the inference step. The total training objective of MoSEs becomes the following:
\begin{equation}
\mathcal{L}_{\text{total}} = \mathcal{L}_{\text{task}} + \lambda_{\text{pull}} \cdot \mathcal{L}_{\text{pull}},
\end{equation}
where $\lambda_{\text{pull}}=0.1$ is a coefficient that balances the influence of the pull constraint.

\Cref{tab:cpk_ablation} evaluates the efficacy of the \textbf{CPK} (\Cref{eq:total_loss}) within the MoSEs framework (\Cref{eq:routing}) on the TRACE benchmark, demonstrating its clear superiority over the standard \textbf{CS} baseline (\Cref{eq:cosine_similarity}) in terms of both performance and stability. While the CS-based approach achieves an accuracy (ACC) of 49.2 with a backward transfer (BWT) of -0.90\%, the CPK method significantly enhances these metrics by leveraging contrastive ($\lambda_1$) and orthogonal ($\lambda_2$) loss components to refine sub-expert routing. As the contrastive coefficient $\lambda_1$ increases to 2.0 and the orthogonality constraint $\lambda_2$ is tuned to 0.4, the model reaches the best performance of 52.8 and a remarkably minimized forgetting rate of -0.50\%. Crucially, this performance gain is achieved with high parameter efficiency, maintaining the 2.15M parameters across all configurations, which proves that the contrastive embedding space in MoSEs provides a more robust and discriminative manifold for task-specific routing without increasing the model's computational complexity.

\begin{table}[t]
\centering
\caption{\small \textbf{Contrasitive Prompt Key (CPK)} of MoSEs on the TRACE benchmark using LLaMA-2-7B.}
\label{tab:cpk_ablation}
\resizebox{0.85\textwidth}{!}{
\begin{tabular}{cccccc|cc|c}
\toprule
\textbf{SubExpert} & \textbf{Sparse Routing} & \textbf{Pull Loss} & \textbf{Pull Coeff.} & \textbf{Cont.} & \textbf{Ortho.} & \textbf{ACC ($\uparrow$)} & \textbf{BWT ($\uparrow$)} & \textbf{\# Params} \\
(\textbf{S}\textbf{E}) & (SR) & (\textbf{CS} or \textbf{CPK}) & $\lambda_{pull}$ & $\lambda_1$ & $\lambda_2$  & \multicolumn{2}{c|}{\textbf{Performance}} & \textbf{Efficiency} \\

\midrule
\checkmark & \checkmark &   - & -   & -      & -    & 42.0    & -3.71\% & 2.15M \\ \midrule
\checkmark & \checkmark & CS  & 0.1 & -      & -    & 49.2    & -0.90\% & 2.15M \\ \midrule

\checkmark & \checkmark & CPK &  -  &  1.0   & 0.3  & 51.8    & -0.80\% & 2.15M \\ 
\checkmark & \checkmark & CPK &  -  &  2.0   & 0.3  & 52.3    & -0.62\% & 2.15M \\
\rowcolor[HTML]{EFEFEF} 
\checkmark & \checkmark & CPK &  -   & 2.0   & 0.4  & \textbf{52.8} & \textbf{-0.50\%} & \textbf{2.15M} \\
\bottomrule
\end{tabular}
}
\vspace{-10.0pt}
\end{table}

\begin{figure}[!h]
    \centering
    \small
    \setlength{\tabcolsep}{0pt}{%
    \begin{tabular}{cc}
    \includegraphics[width=0.5\columnwidth]{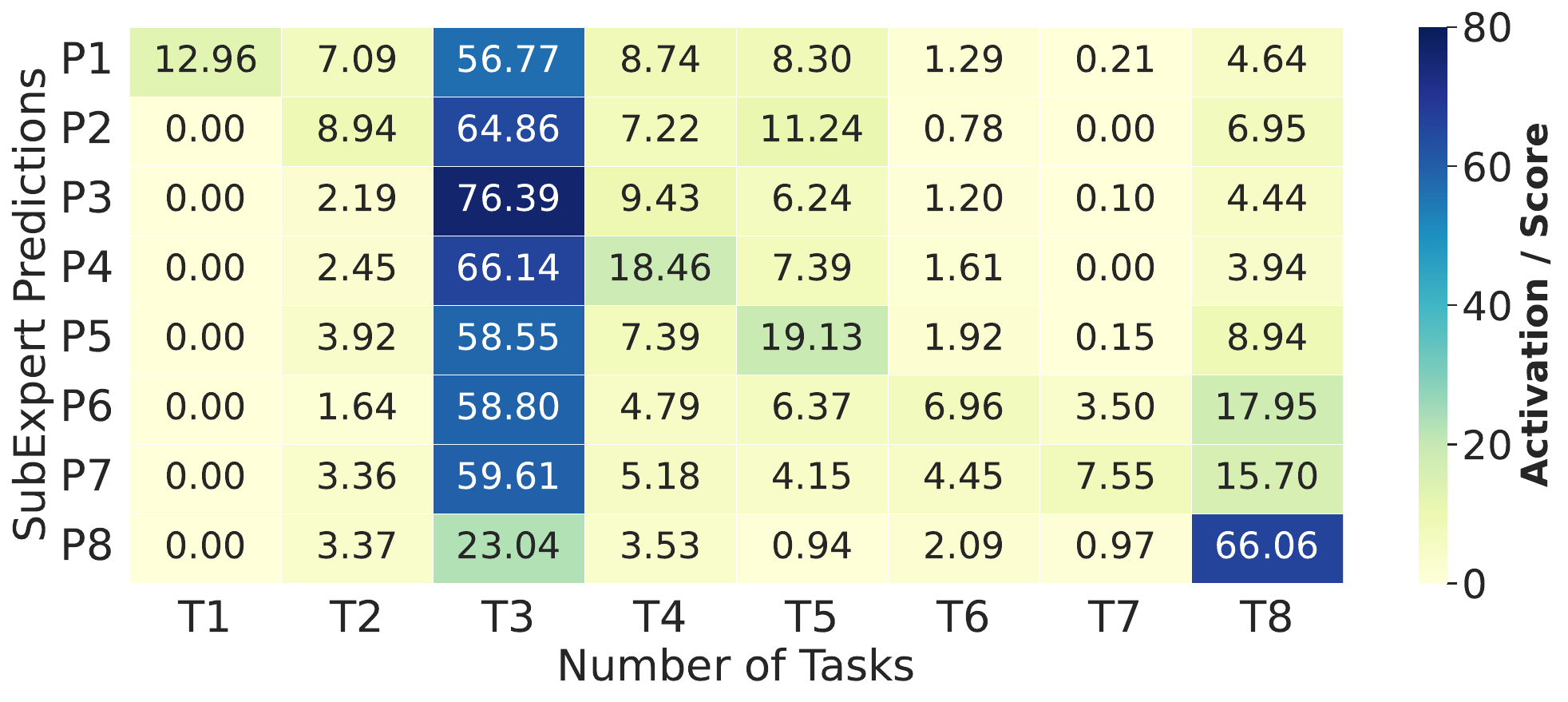} & 
    \includegraphics[width=0.5\columnwidth]{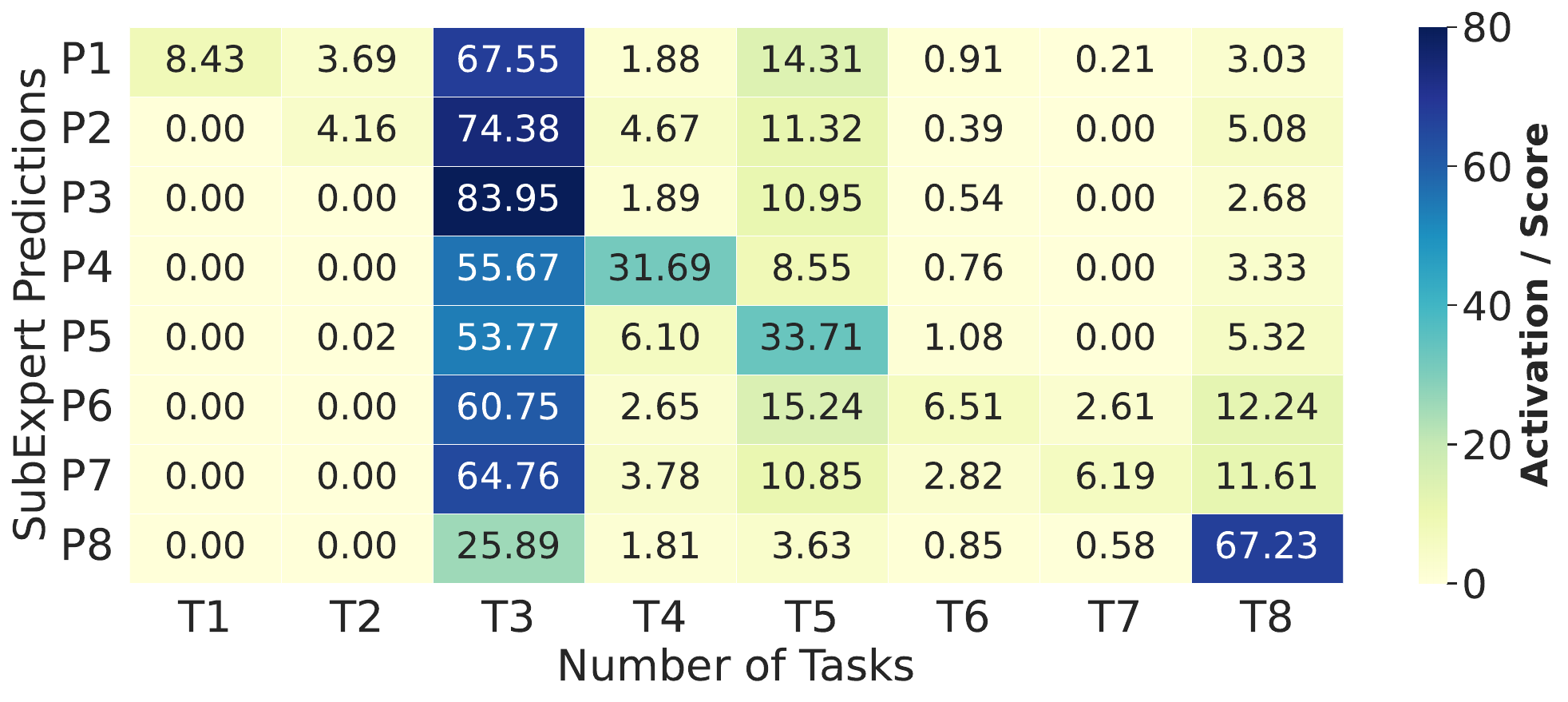} \\
    \vspace{-0.05in}
    \small (a) Cosine Similarity (CS) & \small (b) Contrastive Prompt Key (CPK) Loss \\
    \end{tabular}
    }
    \vspace{-0.05in}
    \caption{\small \textbf{Comparisions of Routing Matrices, MoSEs, $c = 29.0\%$} on the TRACE benchmark dataset.}
    \label{fig:routing_matrix}
    \vspace{-15pt}
\end{figure}

\noindent
The comparison between the two routing matrices demonstrates that incorporating Contrastive Loss \Cref{fig:routing_matrix}(b) significantly enhances task-specific specialization and routing sparsity compared to the standard Cosine Similarity \Cref{fig:routing_matrix}(a) approach. A critical observation occurs at T3, which represents the MeetingBank dataset-a complex task involving lengthy city council transcripts and specialized administrative terminology. While the MeetingBank task (T3) naturally demands high activation across multiple SubExperts due to its linguistic density and reasoning requirements, the Contrastive Loss model achieves a much sharper distribution, peaking at 83.95 (P2). This indicates that the sub-router more effectively isolates the core knowledge required for meeting summarization within dedicated parameters. Furthermore, the Contrastive Loss mechanism minimizes interference in earlier tasks (T1, T2) and secures a highly specialized profile for the final task at T8. These results empirically validate that MoSEs successfully mitigates catastrophic forgetting and optimizes sub-routing even when dealing with high-entropy datasets like MeetingBank (T3), ensuring efficient sublinear capacity growth.

Unlike \textbf{CS}, which merely minimizes the angular distance between a query and a specific key, the \textbf{CPK} supervised by contrastive loss actively reshapes the embedding manifold by enforcing both intra-class compactness and inter-class separability. While CS-based routing often results in a crowded embedding space where task boundaries become blurred, leading to sub-optimal expert selection, contrastive learning explicitly pushes dissimilar task features apart in the latent space. This creates a highly discriminative embedding landscape where each SubExpert's representative key is anchored to a unique, orthogonal region. Consequently, MoSEs (CPK) facilitates a more robust routing decision that effectively isolates task-specific knowledge while maintaining a shared parameter space, thereby significantly reducing catastrophic interference compared to naive similarity-based approaches.

\noindent
\textbf{Broader Impacts.}
The Mixtures of SubExperts (MoSEs) framework provides a highly practical path toward rehearsal-free or rehearsal-light continual adaptation for LLMs, delivering exceptional knowledge retention and operational efficiency. This innovative architecture isolates task-specific knowledge within dedicated SubExperts, maximizing memory stability while achieving SOTA performance through sublinear capacity growth. From a privacy perspective, MoSEs serves as a secure alternative to rehearsal-based methods by storing only parameter artifacts, such as task keys and masks, instead of raw data buffers, thereby mitigating the risks associated with storing sensitive training samples. Furthermore, MoSEs ensures robust scalability over long task horizons by utilizing efficient management policies for pruning and merging keys - similar to the K-D tree strategies that avoid linear complexity - allowing the model to adapt to diverse data distributions effectively. Ultimately, by maintaining an optimal balance between accuracy and resource consumption, MoSEs enhances the reliability and adaptability of innovative AI systems across dynamic real-world domains.

\noindent
\textbf{Limitations \& Future Directions.} While the MoSEs framework effectively minimizes forgetting and achieves superior performance by leveraging shared and isolated parameter structures and contrastive embedding alignments, it is not without limitations that pave the way for future research. Specifically, the routing efficacy still maintains a structural dependency on key prompt embeddings. Furthermore, the selection of sparsity levels and adapter ranks remains a manual hyperparameter tuning process. To address these constraints, future directions will focus on proposing robust adaptive routing mechanisms and dynamic allocation strategies that can autonomously adjust structural parameters in response to shifting domain complexities, thereby further enhancing the scalability and robustness of the MoSEs architecture.

\section{Related Work: Continual Learning (CL)}
Continual learning (CL) aims to enable models to sequentially learn multiple tasks without catastrophic forgetting~\cite{parisi2019continual}. Classical CL methods are broadly categorized into regularization-based, replay-based, and dynamic architecture approaches. Regularization-based methods, such as Elastic Weight Consolidation (EWC)~\cite{kirkpatrick2017overcoming} and Synaptic Intelligence (SI) \cite{zenke2017}, constrain updates to important parameters for past tasks by adding Fisher-based penalties. These methods are effective in small-scale models but degrade under the scale and parameter redundancy of LLMs. Replay-based methods mitigate forgetting by storing a subset of past data for rehearsal~\cite{rebuffi2017icarl, chaudhry2018efficient, chaudhry2019continual, Saha2021, lin2023pcr, liang2024loss}. Although effective, they raise concerns about data privacy and storage costs, particularly in domains such as healthcare and finance. Architecture-based models, such as Progressive Neural Networks~\cite{rusu2016progressive}, Dynamically Expandable Networks~\cite{yoon2018lifelong}, and Supermasks~\cite{wortsman2020supermasks}, allocate new sub-networks~\cite{kang2022forget} for each task or use fixed backbones with binary masks to carve out task-specific subnetworks. While avoiding forgetting, they often lead to linear parameter growth and inference inefficiencies, which are prohibitive in large-scale transformer models. In this work, we designed a novel subexpert routing continual learning method specifically for the computational and memory constraints of LLMs, where retraining or full fine-tuning is costly.

\noindent
\textbf{Public Source Code} All official source codes will be available soon.


\end{document}